\documentclass[letterpaper,journal]{IEEEtran}

\usepackage{amsmath,amsfonts}
\usepackage{array}
\usepackage{booktabs}
\usepackage[caption=false,font=normalsize,labelfont=sf,textfont=sf]{subfig}
\usepackage{textcomp}
\usepackage{url}
\usepackage{graphicx}
\usepackage{xcolor}
\usepackage[tracking=true,protrusion=false,expansion=false]{microtype}

\newcounter{algorithm}
\newcounter{algline}
\newenvironment{algorithm}[1][t]
  {\begin{figure}[#1]\refstepcounter{algorithm}\setcounter{algline}{0}\footnotesize}
  {\end{figure}}
\newcommand{\algorithmcaption}[1]{%
  \centering\textbf{Algorithm~\thealgorithm:} #1\par\vspace{3pt}\raggedright}
\newcommand{\algline}[2][0em]{%
  \stepcounter{algline}\noindent\makebox[1.5em][r]{\thealgline:}\hspace{0.4em}%
  \hspace{#1}#2\par}
\newcommand{\algkw}[1]{\textbf{#1}}
\definecolor{ActorAlg}{HTML}{7A1FA2}
\definecolor{LearnerAlg}{HTML}{D14900}
\newcommand{\algthread}[2]{\algline{\textcolor{#1}{\textbf{#2:}}}}
\usepackage[colorlinks=true,urlcolor=magenta,linkcolor=blue,citecolor=blue]{hyperref}

\let\oldcite=\cite
\renewcommand{\cite}[1]{\textcolor{blue}{\oldcite{#1}}}
\newcommand{\citerange}[3]{\nocite{#1,#2,#3}\cite{#1}\textcolor{blue}{--}\cite{#3}}
\newcommand{\citefiverange}[5]{\nocite{#1,#2,#3,#4,#5}\cite{#1}\textcolor{blue}{--}\cite{#5}}

\begin{document}

\title{%
\textls[10]{VLA-Precision: Asymmetric Co-Bootstrapping for Efficient}\\[-0.1em]
\textls[-10]{Real-World Online RL of Vision-Language-Action Models}}

\author{%
Chenyu Su$^{1}$, Zhaolong Shen$^{3,4}$, Yuan Qian$^{1}$, Chen Qian$^{1}$,
Rui Zhang$^{1}$,\\
Feng Yan$^{1}$, Weixing Chen$^{1}$, Fei Zhang$^{1,5}$, Jiamin Wang$^{1}$,
Shuang Cong$^{1}$, and Weiwei Shang$^{*,1,2}$\\[0.55em]
\begin{minipage}{0.82\textwidth}
\centering\small
$^{1}$University of Science and Technology of China\\[0.12em]
$^{2}$State Key Laboratory of Precision and Intelligent Chemistry\\[0.12em]
$^{3}$Beihang University\qquad
$^{4}$Zhongguancun Academy\qquad
$^{5}$Hefei SpinX Technology\\[0.12em]
$^{*}$Corresponding author: \href{mailto:wwshang@ustc.edu.cn}{\textcolor{black}{wwshang@ustc.edu.cn}}
\end{minipage}%
}

\IEEEaftertitletext{%
\vspace{-3.0\baselineskip}
\begin{center}
    \includegraphics[width=0.95\textwidth]{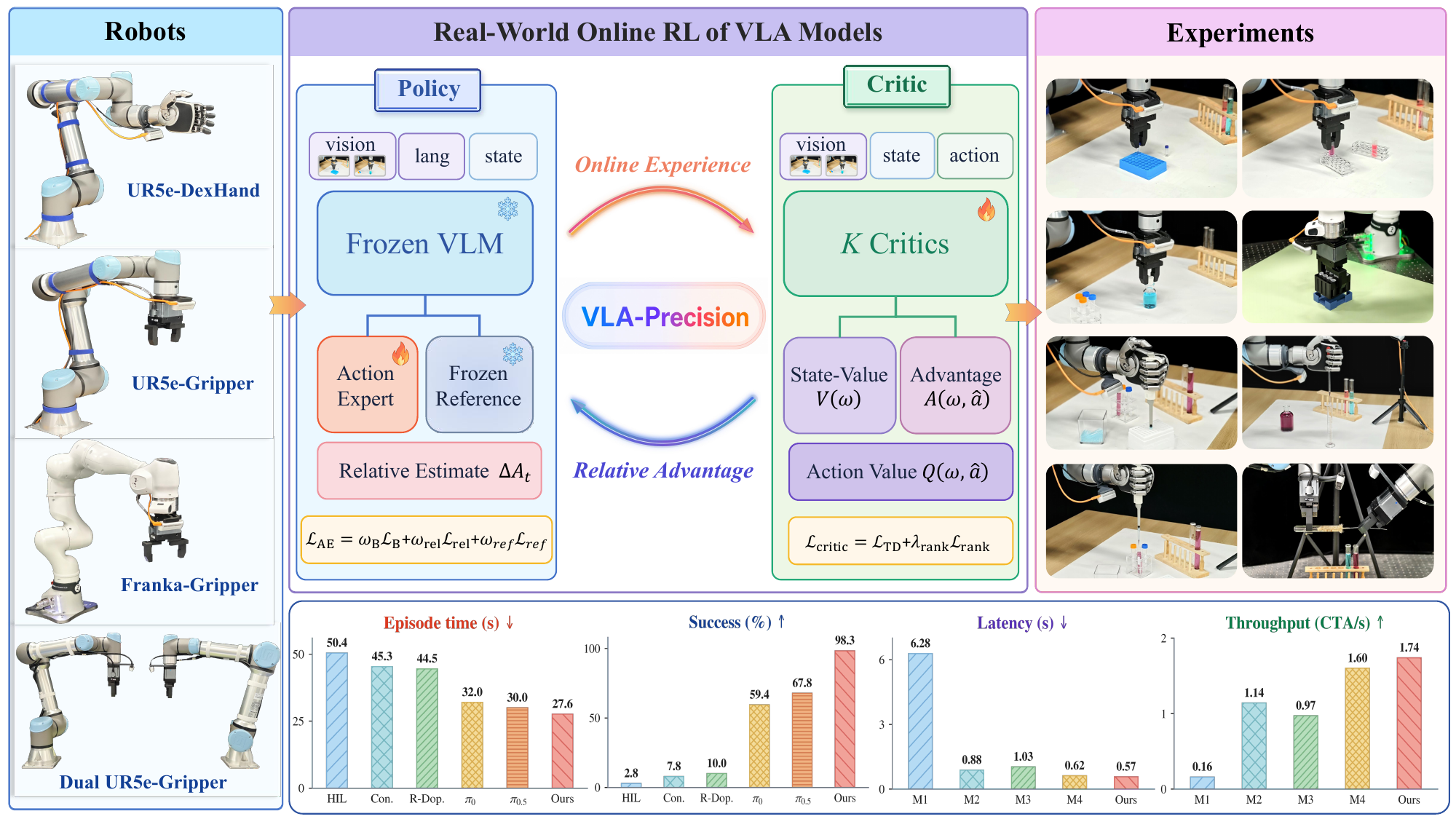}
    \begin{minipage}{0.98\textwidth}
    \expandafter\def\csname @captype\endcsname{figure}
    \caption{Framework overview. VLA-Precision combines rapid behavioral learning, progressive
    value calibration, and reference-regularized relative-advantage policy improvement within \textcolor{black}{an
    efficient training architecture} for \textcolor{black}{real-world online RL of large VLAs}. Across nine high-precision \textcolor{black}{chemistry manipulation tasks} and four
    \textcolor{black}{robotic platforms}, it outperforms baselines in success rate, episode time,
    critic-to-actor latency, and throughput.}
    \label{fig:framework_overview}
    \end{minipage}
\end{center}
\vspace{0.5\baselineskip}
}

\maketitle

\begin{abstract}
Pretrained vision-language-action (VLA) models enable broad manipulation but remain \textcolor{black}{unreliable} in
tasks demanding precision and \textcolor{black}{efficiency}. Applying real-world online reinforcement learning (RL)
to VLA post-training enables autonomous trial-and-error improvement beyond demonstrations alone,
but exposes two bottlenecks: 1) unreliable value signals can induce policy drift; 2) large-VLA
overhead constrains throughput and sample efficiency. To address these challenges, we present
VLA-Precision, an efficient real-world online RL framework featuring the Asymmetric
Co-Bootstrapping (ACoB) algorithm and the ACoB-Stream architecture. Specifically, ACoB
establishes asymmetric co-bootstrapping across timescales: early intervention-guided behavioral
learning rapidly improves policy performance while enhancing online experience quality. As
autonomous experience accumulates, global return propagation and local preference ranking
progressively calibrate value estimates, yielding relative action advantages for reference-regularized
policy improvement while suppressing drift. To enable ACoB on large VLAs, we develop ACoB-Stream,
\textcolor{black}{an efficient online RL training architecture} that establishes invariant-state decoupling and on-demand
streaming as principles, delivering up to 10.9$\times$ improvements in throughput and
computational efficiency. Evaluations on nine high-precision \textcolor{black}{chemistry manipulation tasks} across four
categories and four \textcolor{black}{robotic platforms} show that VLA-Precision achieves 98.3\% mean success rate in
45.8 min/task, with 27.6 s episodes at 1.2$\times$ and 1.8$\times$ the speeds of VLA and RL
baselines. Resources are available at
\href{https://vla-precision.github.io/}{VLA-Precision.github.io}.
\end{abstract}


\section{Introduction}
\IEEEPARstart{R}{ecent} vision-language-action (VLA) models have acquired broadly transferable
capabilities through large-scale pretraining on heterogeneous multimodal
data~\citerange{black2025pi0}{black2025pi05}{kim2025openvla}. With only a few task-specific
demonstrations, VLAs can perform diverse robotic manipulation tasks and generalize
across environmental variations. However, broad adaptability cannot guarantee precise and
\textcolor{black}{efficient} task completion: small residual errors at critical stages still cause intermittent
failures. Behavior cloning (BC) learns task behavior directly from dense action labels in
demonstrations, but errors compound outside the demonstrated distribution and performance remains
bounded by demonstration quality. In contrast, reinforcement learning (RL) uses temporal credit
assignment to learn long-term action values from interaction. Therefore, applying RL to VLA
post-training enables continual policy optimization through interaction-derived return signals,
thereby \textcolor{black}{mitigating compounding errors for improved precision and efficiency, while surpassing demonstration-based performance limits}.

Toward this goal, existing work has explored simulation-based and real-world RL for VLA
post-training~\citerange{guo2025irevla}{xiao2025pld}{zhu2026wmpo}. Simulation-based RL enables massively parallel and failure-tolerant interaction,
supporting scalable and low-cost policy optimization without the physical constraints of real
robots. However, the sim-to-real gap in visual observations and contact dynamics limits reliable
transfer to physical robots, \textcolor{black}{causing alignment errors and unstable contact in precision manipulation}~\cite{lu2025vlarl,zang2025rlinf}. In contrast, real-world RL optimizes
VLAs through direct physical interaction, \textcolor{black}{enabling policies to adapt alignment and contact behaviors through physical feedback}. Existing real-world RL methods fall into two groups by model scale: RL for large
pretrained VLAs~\citerange{physicalintelligence2025pi06}{dong2026expoft}{xu2026rltoken} and RL for
compact robot policies~\citerange{luo2024hilserl}{lei2025rl100}{chen2025conrft}. Despite these
advances, real-world VLA-RL remains constrained by two critical bottlenecks: 1) Algorithmic
stability: unreliable value estimates can induce policy drift; and 2) System efficiency: large-VLA
computational overhead limits training throughput and reduces overall online RL efficiency.

To mitigate the algorithmic bottleneck,
methods~\citerange{chen2025conrft}{zhao2025silri}{shi2026rlco} regularize RL updates with
behavioral-cloning constraints from demonstrations. While these constraints
preserve prior behaviors and stabilize exploration, they provide limited improvement in value
estimation. Consequently, RL updates remain vulnerable to value errors and policy drift\textcolor{black}{, hindering efficient learning to perform tight-tolerance precise manipulation}. For the
systems bottleneck, existing methods~\cite{luo2024hilserl,chen2025conrft,zhao2025silri} rely
on replay-intensive optimization, whose computational cost increases with large VLAs,
limiting training throughput. Real-world online interaction further amplifies this computational
burden. These limitations highlight the need for reliable RL algorithms and efficient \textcolor{black}{online} training \textcolor{black}{architectures for precise manipulation by large VLAs}.

To address the first problem, we propose Asymmetric Co-Bootstrapping (ACoB), a real-world online RL
algorithm that couples rapid behavioral learning with progressive value calibration across
timescales. On the faster timescale, intervention-guided behavioral learning rapidly incorporates
successful and human-executed actions, accelerating policy improvement while continually improving
online experience quality. Meanwhile, global return propagation and local preference ranking
continuously calibrate value estimates as autonomous experience accumulates. The resulting relative
action advantages guide reference-regularized policy improvement, suppressing policy drift while
preserving learned task-specific competence. These asymmetric interactions establish a
cross-timescale co-bootstrapping loop: rapid behavioral learning continuously improves the
experience underlying value estimation, while \textcolor{black}{progressively calibrated value estimates guide policy improvement for precise and efficient manipulation}.

To address the second problem, we propose ACoB-Stream, \textcolor{black}{an efficient training architecture for real-world online RL of large VLAs}. Guided by invariant-state decoupling and on-demand streaming,
ACoB-Stream jointly manages the lifecycles of experience-context state and policy state across four
stages: (1) \textbf{Experience-context formation:} by retaining frozen VLM contexts across updates,
it reduces redundant computation during policy updates; (2) \textbf{Experience-context
persistence:} through context deduplication and sliding-window sampling, it reduces storage
redundancy and limits random-access overhead as experience accumulates; (3)
\textbf{Experience-context access:} through objective-aligned context retrieval, it improves
data-access efficiency during optimization; and (4) \textbf{Policy-state synchronization:} through
trainable-subspace policy dissemination, it reduces synchronization overhead and accelerates policy
deployment. Together, these stages form a closed-loop experience--policy stream, reducing
end-to-end system overhead in real-world online RL under constrained compute and memory.

The contributions of this work are summarized as follows:
\begin{enumerate}
    \item We propose Asymmetric Co-Bootstrapping (ACoB), a real-world online RL algorithm for VLA
	post-training that couples rapid behavioral learning with progressive value calibration
	across timescales, \textcolor{black}{suppressing policy drift induced by value-estimation errors to support precise and efficient real-world manipulation}.
    \item We develop ACoB-Stream, \textcolor{black}{an efficient online RL training architecture} organized around
	state lifecycles, with invariant-state decoupling and on-demand streaming as the design
	principles, delivering up to 10.9$\times$ improvements in throughput and \textcolor{black}{computational} efficiency.
    \item Through extensive evaluations on nine high-precision \textcolor{black}{chemistry manipulation tasks} across four
	categories and four \textcolor{black}{robotic platforms}, VLA-Precision achieves 98.3\% mean success in
	45.8 min/task, with 27.6 s episodes at 1.2$\times$ and 1.8$\times$ the speeds of
	VLA and RL baselines.
\end{enumerate}

The remainder of this paper is organized as follows: In Section \textcolor{blue}{II}, we review
related work on RL-based VLA post-training, including learning algorithms and system architectures.
In Section \textcolor{blue}{III}, we present the proposed ACoB algorithm and ACoB-Stream system
architecture in detail. In Section \textcolor{blue}{IV}, we evaluate the algorithmic effectiveness
and system efficiency of VLA-Precision through extensive real-world experiments on high-precision
\textcolor{black}{chemistry manipulation tasks}. In Section \textcolor{blue}{V}, we discuss the limitations and future directions
of this work. Finally, we conclude this paper in Section \textcolor{blue}{VI}.

\section{Related Work}

\subsection{Simulation-Based RL for VLA Post-Training}
Reinforcement learning (RL) may break demonstration-imposed performance ceilings in
vision--language--action (VLA) models through reward-driven exploration and temporal credit
assignment. Simulation-based RL enables efficient, high-throughput VLA exploration through
large-scale parallel interaction without real-robot training costs and safety risks
\citerange{ma2026vlasurvey}{makoviychuk2021isaacgym}{gu2023maniskill2}. Among the first studies,
RIPT-VLA \cite{tan2025ript} pairs dynamic rollout sampling with leave-one-out advantages, raising
one-shot success from 4\% to 97\%, while VLA-RL \cite{lu2025vlarl} combines trajectory-level RL
with process rewards to outperform fine-tuning by 4.5 points. A controlled VLA benchmark
\cite{liu2025rlgeneralization} shows that RL chiefly strengthens semantic generalization and
execution robustness. Subsequent work broadens task and policy coverage. SimpleVLA-RL
\cite{li2025simplevlarl} combines VLA-specific sampling and parallel rendering, lifting LIBERO-Long
\cite{liu2023libero} from 17.1\% to 91.7\%, while RLinf-VLA \cite{zang2025rlinf} coordinates
heterogeneous workloads for up to $2.27\times$ speedup. For flow-based policies,
$\pi_{\mathrm{RL}}$ \cite{chen2025pirl} constructs tractable likelihoods, improving success by up
to 31.0\%. Recent work targets physical transfer. RL-Co \cite{shi2026rlco} regularizes
simulation-based RL with real demonstrations, raising real-world success by 24\% and 20\%, while
WoVR \cite{jiang2026wovr} controls imagined dynamics to improve success across robot platforms.
Although these methods reduce real-data demands, the sim-to-real gap still limits transfer
reliability. Real-world RL grounds autonomous trial-and-error in real observations and dynamics,
converting pretrained VLA competence into precise, \textcolor{black}{efficient} execution. This advantage
motivates our focus on real-world VLA post-training.

\subsection{Real-World RL for VLA Post-Training}
By learning from physical interaction, real-world RL adapts policies to actual sensing and
dynamics, improving \textcolor{black}{task performance} under deployment conditions. RL for compact robot policies has
advanced contact-rich and dynamic manipulation
\citefiverange{kalashnikov2018qtopt}{zheng2022pageflipping}{zhao2025thindeformable}{aslam2025dartbot}{liu2025constraintmanifold}.
HIL-SERL \cite{luo2024hilserl} reaches near-perfect success in 1-2.5 hours via intervention-guided
online RL, but has limited within-task generalization. ConRFT \cite{chen2025conrft} reaches 96.3\%
mean success in 45--90 min with offline-to-online RL on Octo representations~\cite{ghosh2024octo}, but Q maximization
sacrifices prior performance and permits policy drift. Robo-Dopamine \cite{tan2025robodopamine}
equips ConRFT with a large progress-based reward model, but OOD hallucinations can misguide
optimization. RL-100 \cite{lei2025rl100} applies offline-to-online RL to diffusion policies, but
converges slowly, requiring 12.1 hours of real-robot rollouts per task on average, with 6.8 hours offline and 5.3 hours online. These tradeoffs motivate RL
for large pretrained VLAs, which retain broader action priors. $\pi^{*}_{0.6}$
\cite{physicalintelligence2025pi06} exceeds 90\% success on most tasks through batchwise value
labeling and advantage-conditioned retraining, but requires over 1,000 real-robot rollouts per task on average, severely limiting feedback efficiency. RL
Token \cite{xu2026rltoken} accelerates critical-phase throughput by up to $3\times$, but requires 400–1,000 online episodes per task, while externalized improvements and policy handoffs limit end-to-end adaptation. Distinct from prior methods, ACoB
builds on large pretrained VLAs and couples rapid behavioral learning with progressive value
calibration to suppress policy drift, with ACoB-Stream developed to establish an efficient
continuous real-world online RL loop.

\subsection{Efficient Online RL Systems for VLA Models}
Efficient online RL must shorten the closed-loop experience--policy cycle. Simulation-based VLA-RL systems
batch multimodal decoding, parallelize rendering, and coordinate heterogeneous workloads,
achieving up to $2.27\times$ speedup
\cite{lu2025vlarl,zang2025rlinf,li2025simplevlarl}. However, massively parallel, rapidly
resettable simulators do not transfer directly to real-world online RL. For compact policies,
asynchronous actor--learner systems overlap physical interaction with replay optimization to improve
throughput \cite{luo2024hilserl,luo2024serl,chen2025conrft}. This efficiency does not extend to
large pretrained VLAs because optimization at scale imposes greater computational, state-management,
and synchronization costs. EXPO-FT \cite{dong2026expoft} and RL Token
\cite{xu2026rltoken} avoid propagating RL gradients through the VLA by optimizing residual action
edits and a lightweight token-conditioned policy, respectively. This confines reward-driven
improvements to auxiliary policies, requiring separate editors or switching logic and preventing
their accumulation within a unified policy as tasks expand. SOP \cite{pan2026sop} and USER
\cite{zang2026user} scale real-world online learning through fleet orchestration and heterogeneous
cloud--edge infrastructure, respectively. Scale-out increases throughput with robots and
accelerators without reducing per-update large-VLA computation. Departing from the methods above,
we develop ACoB-Stream, \textcolor{black}{an efficient training architecture} organized around \textcolor{black}{experience-context and policy} state
lifecycles, with invariant-state decoupling and on-demand streaming as the core design principles,
enabling \textcolor{black}{efficient real-world online RL of large VLAs}.

\section{Method}

In this section, we first formulate real-world RL post-training of large VLAs under an asynchronous
actor--learner process as a closed-loop experience--policy optimization problem.
Subsequently, we present the Asymmetric Co-Bootstrapping (ACoB) algorithm to \textcolor{black}{improve policy performance while suppressing value-induced policy drift}.
Thereafter, we develop ACoB-Stream, which establishes a closed-loop
experience--policy stream for large VLAs through experience-context formation, persistence,
access, and policy-state synchronization. Finally, we specify the end-to-end online training and
deployment protocol for continuous real-world post-training of large VLAs.

\begin{figure*}[t]
    \centering
    \includegraphics[width=0.95\textwidth]{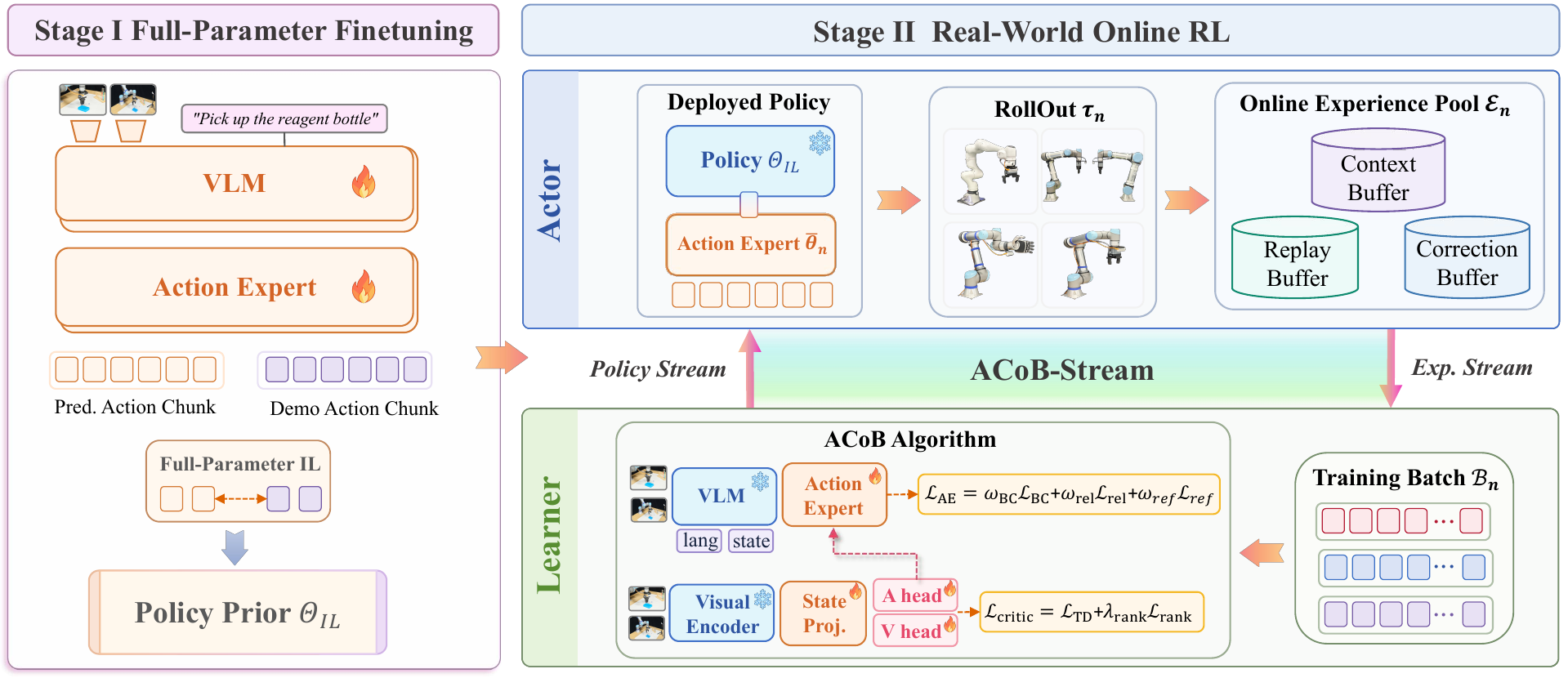}
    \caption{Two-stage post-training pipeline of VLA-Precision. In Stage I, full-parameter imitation
    learning on demonstrations establishes the task-specific policy prior
    $\Theta_{\mathrm{IL}}$. Initialized from $\Theta_{\mathrm{IL}}$, real-world online RL in Stage II
    is formulated as a closed-loop experience--policy optimization problem under an asynchronous
    actor--learner process, where rollouts generate online experience and learner optimization
    yields action-expert updates for redeployment.}
    \label{fig:method_overview}
\end{figure*}

\subsection{VLA-Precision Problem Formulation}

Initializing real-world online RL with demonstrations provides task-relevant priors that
reduce the burden of early exploration and improve sample efficiency~\cite{luo2024hilserl,chen2025conrft}. For a large pretrained VLA,
initialization must additionally establish task-specific competence while preserving
pretrained manipulation priors and within-task generalization.

To this end, Fig.~\ref{fig:framework_overview} provides an overview of the VLA-Precision framework, while
Fig.~\ref{fig:method_overview} details the two-stage post-training pipeline: Stage I performs
full-parameter imitation learning on task demonstrations to obtain the task-specific policy prior
\(\Theta_{\mathrm{IL}}\). Stage II then initializes real-world online RL from
\(\Theta_{\mathrm{IL}}\) and improves the action expert
within this asynchronous actor--learner system. We formulate Stage II as a closed-loop
experience--policy optimization problem.

Specifically, we model physical interaction under task instruction \(\ell\) as a standard Markov
decision process (MDP),
\begin{equation}
\mathcal M_\ell=
\left(\mathcal S,\mathcal A,P,r,\rho_0,\bar\gamma\right).
\label{eq:chunk_mdp}
\end{equation}
Here, \(\mathcal S\) and \(\mathcal A\) denote the state and action spaces, \(P\) \textcolor{black}{denotes} the environment
transition probability, \(r\) \textcolor{black}{is} the reward function, \(\rho_0\) \textcolor{black}{denotes} the initial-state
distribution, and \(\bar\gamma\) \textcolor{black}{is} the discount factor. At decision step \(t\), the state
\(s_t=(o_t,q_t,\ell)\in\mathcal S\) comprises visual observation \(o_t\), robot state \(q_t\), and
task instruction \(\ell\). A VLA policy action is an \(H\)-step action chunk
\(a_t=(u_{t,0},\ldots,u_{t,H-1})\in\mathcal A\subseteq\mathbb R^{H\times d}\), where
\(H\) is the chunk horizon and \(d\) \textcolor{black}{is} the single-step action dimension. Executing the chunk induces
\(s_{t+1}\sim P(\cdot\mid s_t,a_t)\) and yields the cumulative reward
\(R_t^{(H)}=\sum_{h=0}^{H-1}\gamma^h r_{t,h}\), where \(\gamma\) is the primitive-step discount
factor and \(\bar\gamma=\gamma^H\). The task-conditioned policy \(\pi(\cdot\mid s_t)\) maps
each state to an action-chunk distribution and induces trajectory \(\tau\) under \(\mathcal M_\ell\).

To instantiate the task-conditioned policy \(\pi\), we initialize the online policy from the
task-finetuned \(\pi_{0.5}\) and optimize only the LoRA parameters~\cite{hu2022lora} in the action expert. Under this
parameterization, \(\Theta_f\) denotes the frozen multimodal prefix parameters, \(\psi\) \textcolor{black}{denotes} the frozen
action-expert parameters inherited from Stage I, and \(\theta\in\mathbb R^p\) \textcolor{black}{denotes} the trainable LoRA
parameters in the action expert. Action-chunk generation is compactly written as
\begin{equation}
a_t=G_{\psi,\theta}(z_t,\epsilon_t)
\sim\pi_{\Theta_f,\psi,\theta}(\cdot\mid s_t),
\quad \epsilon_t\sim\mathcal N(0,I).
\label{eq:vla_action_expert}
\end{equation}
Here, \(z_t=F_{\Theta_f}(o_t,\ell,q_t)\) is the task-conditioned multimodal prefix context encoded from the visual observation,
task instruction, and robot state. \(\epsilon_t\) is the initial Gaussian noise, \(I\) \textcolor{black}{is} the identity
covariance, and \(G_{\psi,\theta}\) \textcolor{black}{is} the flow-based action expert. The corresponding online RL objective is
\begin{equation}
\begin{gathered}
\theta^\star
=\arg\max_{\theta\in\mathbb R^p}J(\theta),\\
J(\theta)
=\mathbb E_{\tau\sim p(\tau\mid\pi_{\Theta_f,\psi,\theta},P,\rho_0)}
\left[\sum_{t=0}^{T-1}\bar\gamma^t R_t^{(H)}\right].
\end{gathered}
\label{eq:online_objective}
\end{equation}
Here, \(\theta^\star\) denotes the return-maximizing parameters, and \(T\) is the episode horizon
measured in action-chunk decision steps.
To solve Eq.~\eqref{eq:online_objective} through continuous real-world interaction, VLA-Precision
constructs the closed-loop solution process illustrated in Fig.~\ref{fig:method_overview}:
\begin{equation}
\begin{gathered}
\text{Actor:}\;
\bar\theta_n\xrightarrow{\mathrm{Rollout}}\tau_n
\xrightarrow{\mathrm{experience}}\mathcal E_n\ni\mathcal B_n,\\
\text{Learner:}\;
(\theta_n,\mathcal B_n)\xrightarrow{\mathrm{ACoB}}
\theta_{n+1}
\xrightarrow{\mathrm{policy}}\bar\theta_{n+1}
\end{gathered}
\label{eq:closed_loop_cycle}
\end{equation}
Here, \(n\) denotes the online update cycle. \(\theta_n\) and \(\bar\theta_n\) denote the trainable
action-expert state in the learner process and its deployed counterpart in the actor process,
respectively, while \(\tau_n\), \(\mathcal E_n\), and \(\mathcal B_n\) denote the real-world
trajectory, online experience pool, and training batch. The online experience pool comprises the
replay buffer \(\mathcal R\), the correction buffer \(\mathcal C\), and the context buffer
\(\mathcal K\). The actor generates \(\tau_n\) under \(\bar\theta_n\), storing all transitions in
\(\mathcal R\), effective corrections in \(\mathcal C\), and multimodal prefix contexts in
\(\mathcal K\). The learner optimizes \(\theta_n\) with ACoB
over \(\mathcal B_n\subseteq\mathcal E_n\), yielding \(\theta_{n+1}\) for deployment as
\(\bar\theta_{n+1}\). In this formulation, ACoB optimizes the action expert within the learner
process, while ACoB-Stream closes the experience--policy loop between the two processes.

\subsection{Asymmetric Co-Bootstrapping for the Action Expert}

Reliable real-world RL post-training of large VLAs must \textcolor{black}{further improve policy performance through continued interaction}
while preventing unreliable value estimates from inducing policy drift. To this end, ACoB establishes asymmetric co-bootstrapping within the learner process
of Eq.~\eqref{eq:closed_loop_cycle}: behavioral learning rapidly improves the action expert and
continuously enhances online experience quality, while the critic continually calibrates value
estimates as online experience accumulates, yielding increasingly reliable relative action
advantages for further action-expert improvement.

\subsubsection{Progressive Value Calibration}

Reliable value learning requires both long-horizon return estimation and fine-grained credit
assignment. The former captures cumulative behavioral consequences, while the latter sharpens
policy-improvement signals. ACoB therefore establishes progressive value calibration through
global return propagation and local preference ranking for action-expert optimization.

To jointly learn both signals, ACoB instantiates the value model as an ensemble of \(K\) task-specific critics. Each critic adopts
the value--advantage decomposition
\(Q_{\phi_k}(\omega,\widehat a)=V_{\phi_k}(\omega)+
A_{\phi_k}(\omega,\widehat a)\), where
\(\omega_t=(o_t,q_t)=\operatorname{proj}_{o,q}(s_t)\) is the visual--proprioceptive component of
the state and \(Q_{\bar\phi_k}\) is the corresponding target critic. For any action chunk
\(a\in\mathcal A\), let \(\widetilde a=\mathcal T_t(a)\) denote its OpenPI-normalized model representation and
\(\widehat a=\mathcal C(\widetilde a)=\operatorname{vec}(P_c\widetilde a_{0:H})\) its flattened
critic representation, where \(P_c\) excludes inactive action dimensions.

\textit{Global Return Propagation.} Based on the above critic formulation, ACoB propagates long-horizon returns through
temporal-difference optimization. Let \(a_t^{\mathrm{exec}}\in\mathcal A\) denote the action chunk
ultimately executed by the robot, resulting from either autonomous policy inference or human
correction. It induces the transition
\begin{equation}
\begin{gathered}
\xi_t=(s_t,a_t^{\mathrm{exec}},\mathbf r_t,s_{t+1},d_t),\\
\mathbf r_t=(r_{t,0},\ldots,r_{t,H-1}).
\end{gathered}
\label{eq:transition}
\end{equation}
where \(\xi_t\in\mathcal E_n\) and \(d_t\) is the terminal flag. Let
\(\mathcal B_n^{\mathrm{RL}}\subseteq\mathcal B_n\subseteq\mathcal E_n\) denote
samples eligible for value learning. Let \(\widetilde G_{\psi,\theta}\) denote the action-expert
output in the normalized OpenPI action space. For each sampled transition \(\xi_t\), the bootstrap action
at \(s_{t+1}\) and its TD target are constructed as
\begin{equation}
\begin{gathered}
\widehat a_{t+1}^{\theta_n}
=\mathcal C\!\left(\widetilde G_{\psi,\theta_n}(z_{t+1},\epsilon)\right),
\quad \epsilon\sim\mathcal N(0,I),\\
y_t=R_t^{(H)}
+\bar\gamma(1-d_t)\min_k Q_{\bar\phi_k}
(\omega_{t+1},\widehat a_{t+1}^{\theta_n}).
\end{gathered}
\label{eq:td_target}
\end{equation}
Global return propagation is realized by the ensemble TD objective defined as follows:
\begin{equation}
\mathcal L_{\mathrm{TD}}(\phi)
=
\mathbb E_{\mathcal B_n^{\mathrm{RL}}}
\left[
\frac{1}{K}\sum_k
\left(
Q_{\phi_k}(\omega_t,\widehat a_t^{\mathrm{exec}})
-\operatorname{sg}(y_t)
\right)^2
\right],
\label{eq:td_loss}
\end{equation}
where \(\operatorname{sg}\) denotes stop-gradient and
\(\phi=\{\phi_k\}_{k=1}^K\). Through recursive bootstrapping, this objective propagates
long-horizon returns along executed trajectories.

\textit{Local Preference Ranking.} Although corrections can recover a rollout to success, TD over the executed trajectory cannot
correct the value of the overwritten proposal. The proposal may thus remain overvalued and misguide
action-expert optimization. ACoB addresses this limitation through state-matched local preference
ranking, enforcing the correction above the original proposal. Before intervention, the deployed VLA proposes
\(a_t^{\mathrm{prop}}\sim\pi_{\Theta_f,\psi,\bar\theta_n}(\cdot\mid s_t)\). An effective
intervention is marked by \(i_t=1\) and supplies a corrective chunk \(a_t^{\mathrm{hum}}\), giving
\begin{equation}
\begin{gathered}
a_t^{\mathrm{cmd}}=
\begin{cases}
a_t^{\mathrm{prop}}, & i_t=0,\\
a_t^{\mathrm{hum}}, & i_t=1,
\end{cases},\quad
a_t^{\mathrm{exec}}=\operatorname{Exec}(s_t,a_t^{\mathrm{cmd}}).
\end{gathered}
\label{eq:executed_action}
\end{equation}
where \(i_t=0\) denotes autonomous execution and \(\operatorname{Exec}\) denotes environment-side
action resolution. Under intervention, the original proposal is retained for ranking. We set \(c_t=1\)
only when \(i_t=1\) and
\(\lVert\widehat a_t^{\mathrm{exec}}-\widehat a_t^{\mathrm{prop}}\rVert_2>\varepsilon_a\), and
\(c_t=0\) otherwise, where \(\varepsilon_a>0\) is a small tolerance.
Because \(a_t^{\mathrm{prop}}\) has no observed successor, it is excluded from TD learning and used only in the
state-matched ranking objective
\begin{equation}
\begin{gathered}
\mathcal L_{\mathrm{rank}}(\phi)
=
\mathbb E_{\xi_t\sim\mathcal B_n^{\mathrm{RL}}\mid c_t=1}
\left[
\frac{1}{K}\sum_{k=1}^K
\left[m_c-\Delta A^{\mathrm{pair}}_{t,k}\right]_+^2
\right],\\
\Delta A^{\mathrm{pair}}_{t,k}
=
A_{\phi_k}(\omega_t,\widehat a_t^{\mathrm{exec}})
-A_{\phi_k}(\omega_t,\widehat a_t^{\mathrm{prop}}),
\end{gathered}
\label{eq:rank_loss}
\end{equation}
where \(m_c\geq 0\) is the ranking margin and \([x]_+=\max(x,0)\). Together, global return
propagation and local preference ranking define the ACoB critic objective
\begin{equation}
\mathcal L_{\mathrm{critic}}
=
\mathcal L_{\mathrm{TD}}
+\lambda_{\mathrm{rank}}\mathcal L_{\mathrm{rank}}.
\label{eq:critic_loss}
\end{equation}
where \(\lambda_{\mathrm{rank}}\ge 0\) balances the two signals. TD grounds the critic in
long-horizon returns, while local ranking directly calibrates the relative value of the correction
and original proposal, resolving the local credit ambiguity that long-horizon TD propagation cannot
address. Under
the decomposition \(Q=V+A\), \(V\) represents the state-only component shared across actions,
whereas the ranking loss acts only on \(A\). This isolates action-dependent differences from
the state-only component and aligns the signal with action-expert optimization.

\subsubsection{Relative-Advantage Policy Improvement}

Real-world online RL post-training of a large VLA must translate critic value estimates into reliable
action-expert optimization signals. Direct maximization of absolute \(Q\) values can amplify early
value-estimation errors and drive the policy toward spuriously high-value actions. ACoB therefore
proposes relative-advantage policy improvement, replacing absolute-value pursuit with paired
advantage comparison. By focusing on improvement between actions, this mechanism reduces sensitivity
to value scale and optimistic estimation errors.

However, value-signal reliability must be established gradually as online experience
accumulates, so this indirect RL signal alone yields slow, unstable early action-expert updates.
ACoB therefore couples relative-advantage improvement with flow-matching behavior cloning~\cite{lipman2023flowmatching}.
Unlike prior methods that primarily use behavior cloning to constrain the policy
distribution~\cite{chen2025conrft,shi2026rlco}, ACoB uses critical-stage human corrections to drive
behavior cloning early in training, rapidly absorbing high-quality corrections to improve the action
expert and accelerate critic calibration through better experience. Long-horizon return
propagation and local preference ranking then continually calibrate value estimates, providing
increasingly reliable guidance for autonomous improvement as intervention recedes. This interaction
establishes asymmetric co-bootstrapping between rapid behavior cloning and progressive value
calibration. Meanwhile, the task-finetuned action expert is retained as a frozen reference.
Reference regularization \textcolor{black}{limits deviations from the task-finetuned policy},
preserving competence and suppressing value-induced policy drift.

\begin{figure*}[!t]
    \centering
    \includegraphics[width=0.98\textwidth]{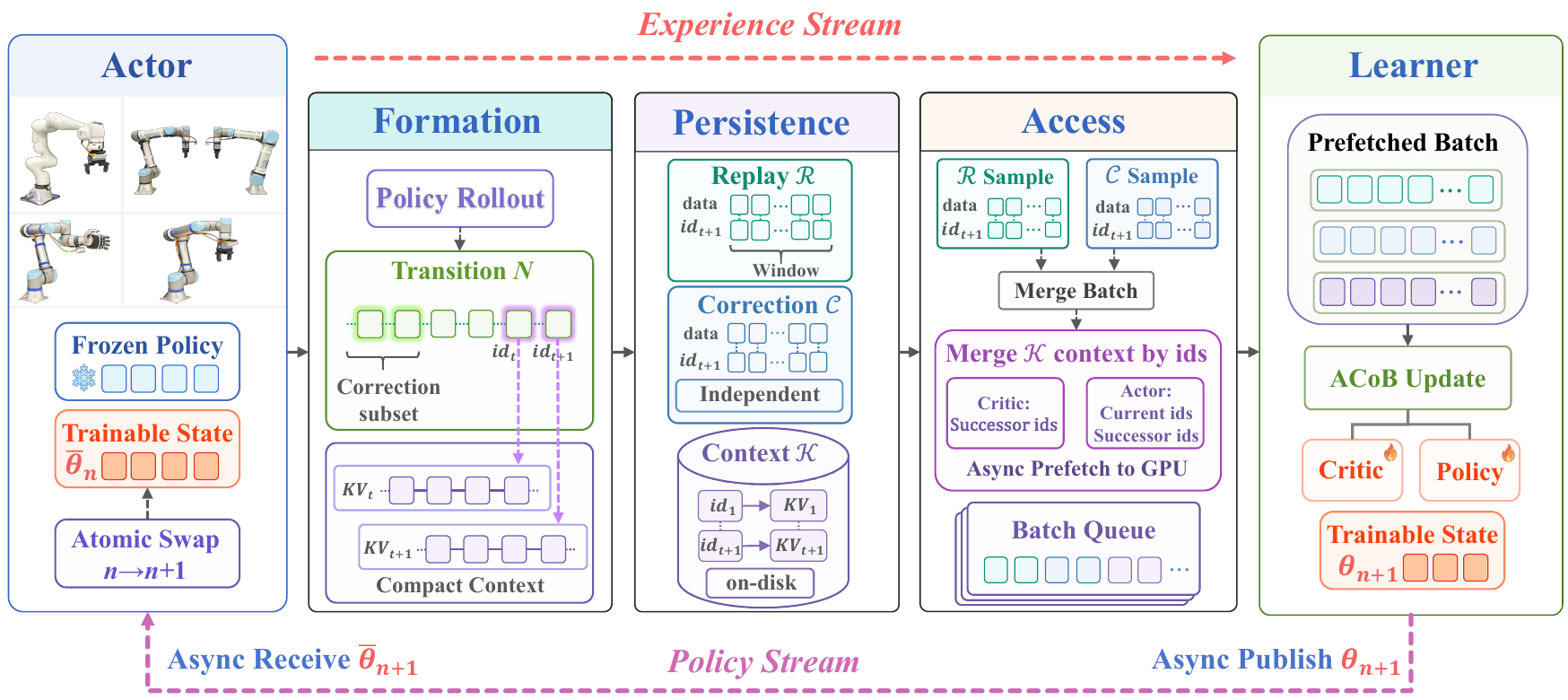}
    \caption{ACoB-Stream architecture. Through invariant-state decoupling and on-demand streaming,
    ACoB-Stream jointly manages the lifecycles of experience-context state and policy state across
    experience-context formation, persistence, objective-aligned access, and policy-state
    synchronization, thereby establishing a closed-loop experience--policy stream between the actor
    and learner.}
    \label{fig:acob_stream}
\end{figure*}

\textit{Relative-Advantage Improvement.} Directly differencing independently aggregated \(Q\) estimates can mix state value, critic-specific
scales, and stochastic decoding variation, obscuring action improvement. ACoB instead computes
paired advantage differences within each critic before pessimistic ensemble aggregation. Given a
multimodal context \(z_t\), the current action expert and frozen reference decode action chunks using the
same random sample \(\epsilon\):
\(\widetilde a_t^\theta=\widetilde G_{\psi,\theta}(z_t,\epsilon)\) and
\(\widetilde a_t^{\mathrm{ref}}=\widetilde G_{\psi,\theta_0}(z_t,\epsilon)\), where \(\theta_0\)
denotes the frozen reference parameters.
Their critic representations are \(\widehat a_t^\theta=\mathcal C(\widetilde a_t^\theta)\) and
\(\widehat a_t^{\mathrm{ref}}=\mathcal C(\widetilde a_t^{\mathrm{ref}})\). For compactness, let
\(A_{t,k}^{x}=A_{\phi_k}(\omega_t,\widehat a_t^{x})\) for
\(x\in\{\theta,\mathrm{ref},\mathrm{prop}\}\). On this basis, ACoB defines the
pessimistic relative advantage as the minimum current-to-baseline advantage gap across the \(K\)
critics:
\begingroup
\setlength{\abovedisplayskip}{1ex plus 3pt minus 1pt}
\setlength{\belowdisplayskip}{1ex plus 3pt minus 1pt}
\begin{equation}
\begin{aligned}
b_{t,k} &= A_{t,k}^{\mathrm{ref}}
+c_t\big[A_{t,k}^{\mathrm{prop}}-A_{t,k}^{\mathrm{ref}}\big]_+,\\
\Delta A_t &= \min_k\left[A_{t,k}^{\theta}-\operatorname{sg}(b_{t,k})\right].
\end{aligned}
\label{eq:relative_advantage}
\end{equation}
\endgroup
where \([x]_+=\max\{x,0\}\). Here, \(c_t\) = 1 for a valid human correction, whereas \(c_t\) = 0
otherwise. The term \(b_{t,k}\) is the comparison baseline of critic \(k\). It equals
\(A_{t,k}^{\mathrm{ref}}\) for \(c_t\) = 0 and
\(\max\{A_{t,k}^{\mathrm{ref}},A_{t,k}^{\mathrm{prop}}\}\) for \(c_t\) = 1. Taking the minimum over
the \(K\) critics makes \(\Delta A_t>0\) only when every critic favors the current action over its
baseline.
To prevent the action expert from pursuing critic scores after sufficient relative improvement,
ACoB optimizes the smooth soft-margin objective
\begin{equation}
\mathcal L_{\mathrm{rel}}(\theta)
=
\mathbb E_{\xi_t\sim\mathcal B_n^{\mathrm{RL}},\epsilon}
\left[
\kappa\operatorname{softplus}\!\left(\frac{m_\pi-\Delta A_t}{\kappa}\right)
\right],
\label{eq:relative_loss}
\end{equation}
where \(\operatorname{softplus}(x)=\log(1+e^x)\), \(m_\pi\geq 0\) is the desired advantage margin,
and \(\kappa>0\) controls its softness.
Critic parameters and comparator branches are held fixed, so gradients reach \(\theta\) only through
\(\widehat a_t^\theta\).

\textit{Flow-Matching Behavior Cloning.} ACoB optimizes flow-matching behavior cloning over task
demonstrations, successful online executions, and effective human corrections. Its flow-matching
objective is
\begin{equation}
\begin{gathered}
\ell_t^{\mathrm{FM}}(\theta)
=\frac{1}{H d_m}\left\|
\left[v_{\psi,\theta}(x_\eta,z_t,\eta)-u_\eta\right]_{0:H}
\right\|_F^2,\\
\mathcal L_{\mathrm{BC}}(\theta)
=\mathbb E_{\xi_t\sim\mathcal B_n,\,\epsilon,\eta}
\left[\ell_t^{\mathrm{FM}}(\theta)\mid\chi_t^{\mathrm{beh}}=1\right].
\end{gathered}
\label{eq:bc_loss}
\end{equation}
where \(v_{\psi,\theta}\) is the velocity field underlying \(\widetilde G_{\psi,\theta}\),
\(\epsilon\sim\mathcal N(0,I)\), \(\eta\sim p_\eta\),
\(x_\eta=\eta\epsilon+(1-\eta)\widetilde a_t^{\mathrm{exec}}\),
\(u_\eta=\epsilon-\widetilde a_t^{\mathrm{exec}}\), and \(d_m\) is the model action dimension.
Here, \(y_t^{\mathrm{suc}}\in\{0,1\}\) denotes the episode-success label of \(\xi_t\).
The behavioral mask \(\chi_t^{\mathrm{beh}}=\max\{y_t^{\mathrm{suc}},c_t\}\) selects all chunks
from successful trajectories and only effective corrections from failed ones.

\textit{Reference Regularization.} To retain the reference behavior during online refinement, ACoB regularizes body-action
coordinates toward the corresponding frozen-reference action:
\begin{equation}
\mathcal L_{\mathrm{ref}}(\theta)
=
\mathbb E_{\xi_t\sim\mathcal B_n,\epsilon}
\left[
\frac{1}{H d_b}\left\|
P_b(\widetilde a_{t,0:H}^\theta-\widetilde a_{t,0:H}^{\mathrm{ref}})
\right\|_F^2
\right].
\label{eq:reference_loss}
\end{equation}
where \(P_b\) selects the body-action coordinates and \(d_b\) is their single-step dimension. The
complete action-expert objective is
\begin{equation}
\mathcal L_{\mathrm{AE}}
=
w_{\mathrm{BC}}\mathcal L_{\mathrm{BC}}
+w_{\mathrm{rel}}\mathcal L_{\mathrm{rel}}
+w_{\mathrm{ref}}\mathcal L_{\mathrm{ref}},
\label{eq:actor_loss}
\end{equation}
where the nonnegative weights sum to one. Here, \(\mathcal L_{\mathrm{rel}}\),
\(\mathcal L_{\mathrm{BC}}\), and \(\mathcal L_{\mathrm{ref}}\) denote relative-advantage RL,
behavior cloning, and reference regularization, respectively.

Together, Eqs.~\eqref{eq:critic_loss} and~\eqref{eq:actor_loss} establish ACoB's asymmetric
co-bootstrapping across timescales: intervention-driven behavior cloning rapidly improves both the
action expert and the experience underlying value calibration, while progressively calibrated value
estimates provide relative-advantage guidance for stable autonomous improvement, with the frozen
reference preserving existing competence.

\subsection{ACoB-Stream Architecture}
Real-world online RL relies on asynchronous actor--learner systems for efficient experience collection
and policy updates. For compact policies, four scale-dependent costs remain manageable: inexpensive
model forward passes, CPU-resident lightweight training samples, low-latency I/O at limited data scale,
and rapid full-model transfer and deployment. By contrast, large VLAs amplify these costs through expensive
frozen-prefix forward, prefix KV accumulation beyond CPU RAM, high-latency large-context access, and
lag-inducing full-model synchronization. Together, these costs prevent new
experience from promptly affecting the deployed policy.

To address this mismatch, ACoB-Stream introduces a closed-loop experience--policy architecture organized
around state lifecycles, jointly managing experience-context state and policy state across four stages:
experience-context formation, persistence, and access, followed by policy-state synchronization. The architecture follows two design principles: invariant-state decoupling avoids repeated
computation, storage, and transfer, while on-demand streaming limits context access and policy
synchronization to the state required by the current objective.
Fig.~\ref{fig:acob_stream} illustrates this four-stage architecture.

\subsubsection{Experience-Context Formation}

Repeated model forward passes through the frozen prefix for every sampled item constitute a primary
throughput bottleneck in large-VLA learner updates. In ACoB-Stream, the experience--context state
comprises the data maintained in the replay buffer \(\mathcal R\), correction buffer \(\mathcal C\),
and context buffer \(\mathcal K\). Because the frozen prefix remains unchanged during online
optimization, ACoB-Stream retains prefix KV caches generated during actor inference. Before
persistence, observation-schema-aware compaction removes inactive prefix positions. At episode
commitment, all transitions enter \(\mathcal R\), their effective-correction subset enters
\(\mathcal C\), and compacted prefix contexts enter \(\mathcal K\). Transition records in
\(\mathcal R\) and \(\mathcal C\) retain current and successor context identifiers that reference
\(\mathcal K\). These identifiers let the learner retrieve the required contexts and train the action
expert without recomputing the frozen prefix. This reduces frozen-prefix computation from once per
sampled item to once per new observation.

\begin{figure}[!t]
    \centering
    \includegraphics[width=\columnwidth]{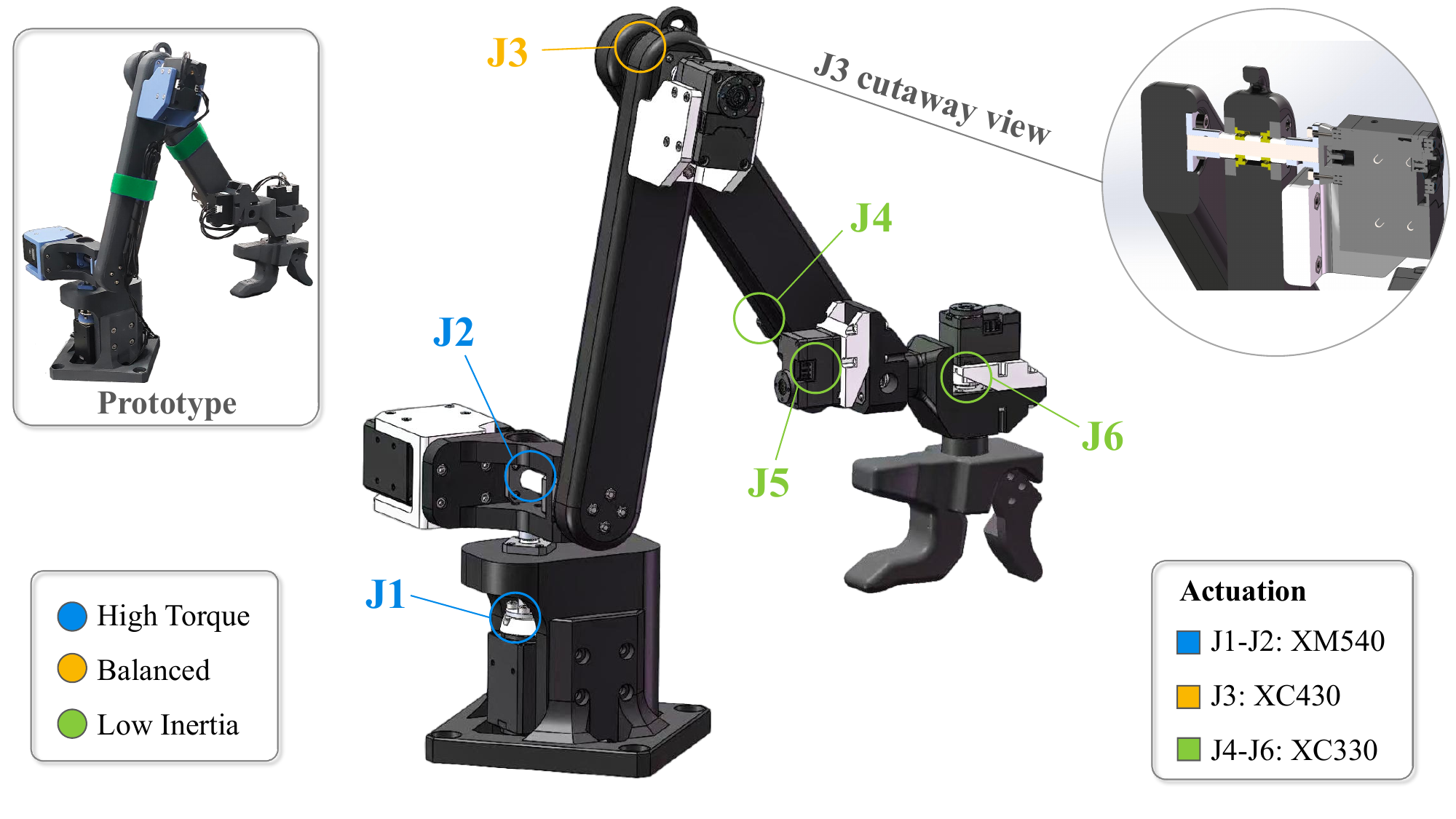}
    \caption{\textcolor{black}{UR5e-isomorphic master arm}: CAD model and prototype.}
    \label{fig:isomorphic_master}
\end{figure}

\subsubsection{Experience-Context Persistence}

Within the experience--context state, the prefix contexts in \(\mathcal K\) pose the primary
persistence challenge: they must remain available across updates, yet their accumulated footprint
outgrows CPU RAM even after compaction. ACoB-Stream therefore designs persistence around context
deduplication and sliding-window sampling. Each cache is stored once in the disk-backed context buffer
\(\mathcal K\). The replay and
correction buffers retain only current and successor context identifiers for all online transitions
and the effective-correction subset, respectively. For replay sampling, the sliding window bounds the active set according to the Linux page-cache capacity available in CPU RAM, improving page-cache reuse and reducing repeated disk-to-CPU data transfers, thereby accelerating data access without requiring explicit KV-cache residency. Correction sampling remains
independent. Unlike a RAM-only sliding buffer, this design retains full history for checkpoint recovery
while reducing duplicate storage and cold-disk access. Sampled transitions and identifiers then pass
to the access stage.

\subsubsection{Experience-Context Access}

Although the disk-backed context buffer \(\mathcal K\) makes accumulated prefix contexts tractable,
random I/O and unnecessary retrieval can still delay learner updates. ACoB-Stream therefore aligns
experience--context access with each optimization objective: value optimization retrieves only
successor contexts from \(\mathcal K\) for its bootstrap targets, whereas action-expert optimization
additionally retrieves current contexts. Samples from \(\mathcal R\) and \(\mathcal C\) are merged
before their required contexts are retrieved in batches from \(\mathcal K\) by identifier. Background
CPU workers sample transitions, retrieve contexts, and assemble host batches, while device workers
transfer the completed batches to GPU and enqueue them after placement. Preparing the next GPU-ready
batch during the current network update reduces learner idle time and improves GPU utilization. The
learner uses these batches for ACoB optimization and passes the updated action-expert state to
synchronization.

\subsubsection{Policy-State Synchronization}

Completing learner optimization affects robot behavior only after the actor process deploys the updated
policy. Full-VLA serialization, transfer, and loading delay deployment and increase policy-version
lag. ACoB-Stream designs synchronization around the trainable policy subspace. \textcolor{black}{Policy updates are published and received asynchronously while training and rollouts continue.} Invariant VLA
state remains resident in both processes. The learner periodically publishes the complete trainable
action-expert state instead of the full VLA or parameter deltas. Upon receipt, the actor merges
this state into its VLA, completes device placement, and atomically replaces the active state
under lock. Rollouts continue on the last valid version during preparation. Transmitting and loading
only changed policy state reduces synchronization and deployment overhead, shortens policy-refresh
latency, and deploys updated policies sooner.

In summary, ACoB-Stream manages state lifecycles through cross-update reuse of frozen-prefix contexts,
deduplicated disk-backed persistence and objective-aligned retrieval, and trainable policy-state
dissemination instead of full-model synchronization.
This reduces computation, memory, I/O, and policy-refresh overhead, allowing new experience to affect
the deployed policy sooner.

\subsection{Real-World Training and Deployment}

Real-world training and deployment of VLA-Precision comprise three components: teleoperation and data
collection, two-stage algorithm training, and robot deployment, which respectively collect real-world
experience, optimize the policy, and execute learned behaviors.

\begin{algorithm}[!t]
\algorithmcaption{Two-Stage Training of VLA-Precision}
\label{alg:acob}
\hrule\vspace{3pt}
\textbf{Require:} pretrained VLA \(\pi_{0.5}\); demonstrations \(\mathcal D\); trainable
action-expert LoRA \(\theta\) and frozen Stage-I reference \(\theta_0\); critic and target ensembles
\(\{Q_{\phi_k},Q_{\bar\phi_k}\}_{k=1}^{K}\); replay buffer \(\mathcal R\), empty correction buffer
\(\mathcal C\), and context buffer \(\mathcal K\); batch size \(B\), update ratio \(N_c\), critic
warm-up steps \(N_w\), and policy publication interval \(N_{\mathrm{pub}}\).
\par\vspace{2pt}
\algline{\textit{Stage I:} Fully fine-tune \(\pi_{0.5}\) on \(\mathcal D\) to obtain \((\Theta_f,\psi)\).}
\algline{\textit{Stage II:} Freeze \((\Theta_f,\psi,\theta_0)\); launch actor and learner; publish \(\theta\).}
\algthread{ActorAlg}{Actor}
\algline[0.8em]{\algkw{for} each episode \(e\) \algkw{do}}
\algline[1.6em]{Set episode trajectory \(\tau_e\leftarrow\emptyset\).}
\algline[1.6em]{\algkw{for} each interaction step \(t\) \algkw{do}}
\algline[2.4em]{\((z_t,a_t^{\mathrm{prop}})\leftarrow
\operatorname{Infer}(\pi_{\Theta_f,\psi,\bar\theta_n},s_t)\).}
\algline[2.4em]{Obtain \(\xi_t=(s_t,a_t^{\mathrm{exec}},\mathbf r_t,s_{t+1},d_t)\) from
Eqs.~\eqref{eq:transition} and~\eqref{eq:executed_action}.}
\algline[2.4em]{Append \((\xi_t,a_t^{\mathrm{prop}},c_t)\) to \(\tau_e\).}
\algline[1.6em]{\algkw{end for}}
\algline[1.6em]{Attach \(y_e^{\mathrm{suc}}\) to all transitions in \(\tau_e\).}
\algline[1.6em]{Append \(\tau_e\), \(\tau_e[c_t=1]\), and \(\{z_t\}\) to \(\mathcal R\), \(\mathcal C\), and
\(\mathcal K\), respectively.}
\algline[0.8em]{\algkw{end for}}
\algthread{LearnerAlg}{Learner}
\algline[0.8em]{\algkw{wait until} \(|\mathcal R|\geq B\) and \(|\mathcal C|\geq B\).}
\algline[0.8em]{\algkw{for} \(w=1,\ldots,N_w\) \algkw{do}}
\algline[1.6em]{Sample \(B/2\) from each of \(\mathcal R\) and \(\mathcal C\), then retrieve contexts from
\(\mathcal K\).}
\algline[1.6em]{Update critics \(N_c\) times by Eq.~\eqref{eq:critic_loss}.}
\algline[0.8em]{\algkw{end for}}
\algline[0.8em]{\algkw{for} \(n=1,2,\ldots\) \algkw{do}}
\algline[1.6em]{Sample \(B/2\) from each of \(\mathcal R\) and \(\mathcal C\), then retrieve contexts from
\(\mathcal K\).}
\algline[1.6em]{Update critics \(N_c\) times by Eq.~\eqref{eq:critic_loss} and \(\theta\) once by
Eq.~\eqref{eq:actor_loss}.}
\algline[1.6em]{\algkw{if} \(n\bmod N_{\mathrm{pub}}=0\) \algkw{then}}
\algline[2.4em]{Publish the updated \(\theta\) to the actor.}
\algline[1.6em]{\algkw{end if}}
\algline[0.8em]{\algkw{end for}}
\vspace{3pt}\hrule
\end{algorithm}

\subsubsection{Teleoperation and Data Collection}

Task requirements for motion range, adjustment precision, and human--machine compliance motivate two
distinct teleoperation schemes: isomorphic master-based teleoperation for long sequences and
compliant interaction, and keyboard-based teleoperation for tasks with limited orientation
variation and fine adjustment.

\textit{\textcolor{black}{Isomorphic Master Arm Interface.}} For the UR5e platform, we design and build the
six-DoF kinematically isomorphic master arm interface shown in
Fig.~\ref{fig:isomorphic_master}. Its joint topology and axes match the UR5e, while parameterized
links support an arbitrary task-specific scale factor and preserve direct joint-space mapping
without online inverse kinematics. The structural parts can be 3D-printed directly from the design
files and assembled with off-the-shelf components, enabling straightforward replication. To
preserve the device and action distribution across demonstration collection and online correction,
the master follows the slave and maintains configuration alignment during policy execution, then
maps operator input to the slave for immediate takeover upon intervention.

This \textcolor{black}{isomorphic master arm} poses two hardware challenges \textcolor{black}{in practical teleoperation}. First, insufficient actuation prevents the master
from following policy-driven slave motion, whereas high manual resistance impedes intervention,
disrupting the transition from policy execution to operator takeover. Second, transmitting link
loads through the servos increases drivetrain backlash and gear wear, degrading joint-mapping
accuracy and long-term reliability. In GELLO~\cite{wu2024gello}, uniform low-torque servos cannot
reliably drive the master to follow policy-controlled slave motion, while servo-borne link loads
accumulate backlash and gear wear under repeated operation. In contrast, we address the former with
load-aware graded actuation and the latter with a load-decoupled joint structure.

In the graded actuation design, we match servo torque and mass to the load at each joint. Specifically, high-torque
XM540 servos at J1--J2 carry the dominant gravity loads, an XC430-T240BB balances torque and mass
at J3, and lightweight XC330-T288 servos at J4--J6 reduce wrist inertia. Accordingly, the proximal
joints provide the torque required for policy following and gravity compensation while distal
inertia and manual resistance remain limited.

However, torque alone cannot ensure structural stability over prolonged data collection.
We therefore use an independent shaft, dual-bearing support, and coupling to decouple load bearing
from actuation. In this structure, the support carries radial and bending
loads, while the servo provides only actuation. Consequently, the resulting load path limits
drivetrain backlash and gear wear while preserving joint stiffness and transmission consistency
during long-term operation.

\begin{figure*}[t]
    \centering
    \includegraphics[width=\textwidth]{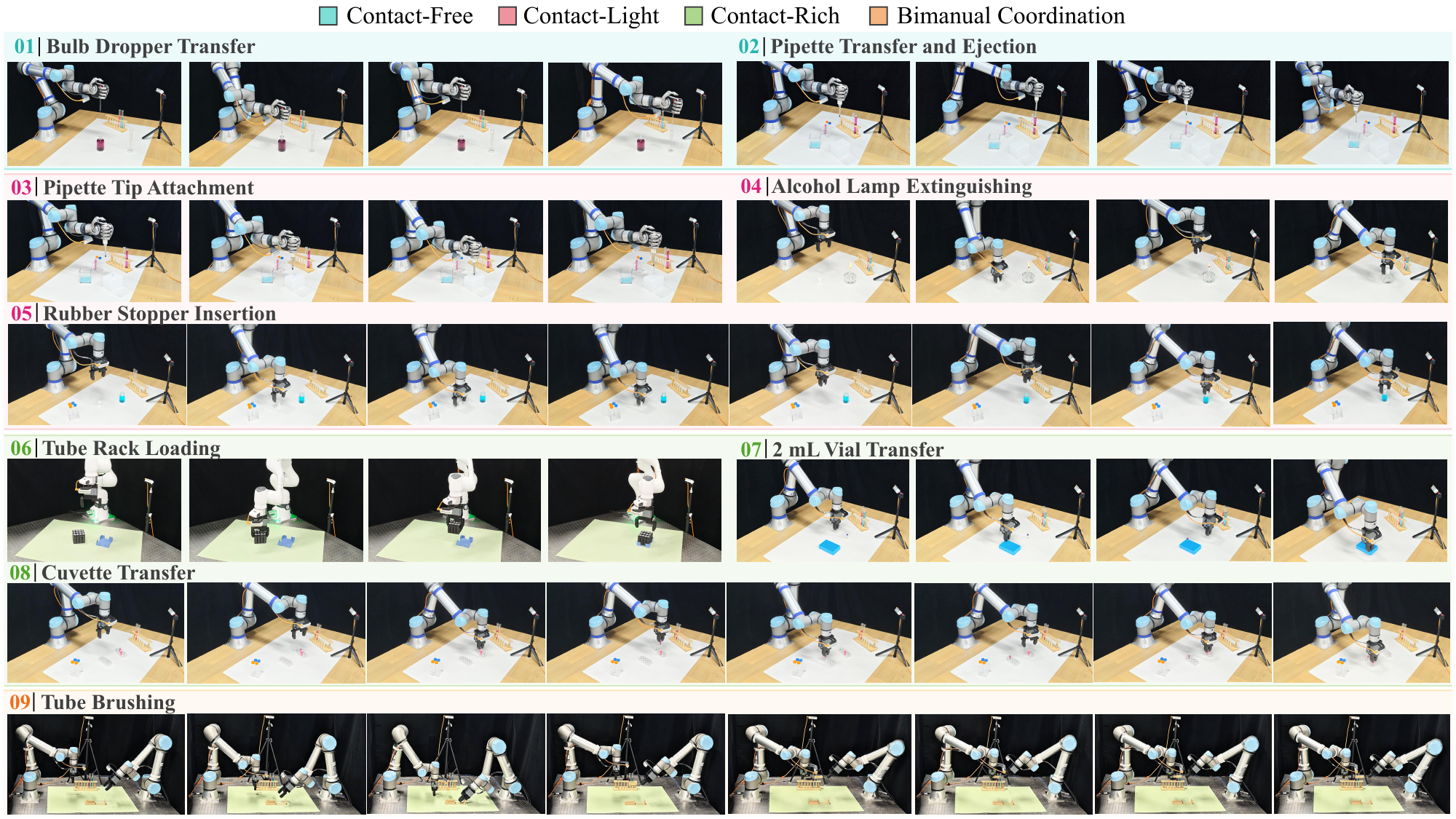}
    \caption{Real-world chemistry manipulation task suite.}
    \label{fig:additional_realworld_examples}
\end{figure*}

\textit{Incremental Cartesian Keyboard Interface.} For stages with limited orientation variation but high
precision demands, continuous teleoperation interfaces, including VR controllers and isomorphic
masters, are ill-suited to sustained alignment at millimeter and submillimeter scales because hand
tremor, device noise, and mapping error directly perturb control commands. To address this limitation,
our keyboard interface maps keystrokes to predefined six-DoF translational or rotational increments
in a fixed task frame and provides coarse and fine step sizes.
Coarse steps accelerate target approach, whereas fine steps provide repeatable corrections at
millimeter or submillimeter scales without requiring the operator to hold a precise pose. This discrete
control improves precision and repeatability during final alignment. However, repeated keystrokes
become cumbersome for long sequences or large reorientations, for which the isomorphic interface is
preferred.

\textit{State--Action Representation.} We adopt a step-wise delta task-space action representation, whose
bounded single-step end-effector increments facilitate stable optimization and fine-motion learning.
Under this representation, the robot state and action are recorded as
\begin{equation}
\begin{gathered}
q_t=[x_{0\rightarrow t}^{\mathrm{rel}},\dot x_t,f_t,\mu_t,g_t]\in\mathbb R^{19},\\
u_t=[\delta x_t^{\mathrm{cmd}},g_t^{\mathrm{cmd}}]\in\mathbb R^7,
\end{gathered}
\label{eq:data_representation}
\end{equation}
where \(x_{0\rightarrow t}^{\mathrm{rel}}\in\mathbb R^6\) represents the \textcolor{black}{tool center point (TCP)} pose relative to the episode start,
\(\dot x_t\in\mathbb R^6\) is the TCP twist, \(f_t,\mu_t\in\mathbb R^3\) are measured force and
torque, which provide critical contact feedback for precise manipulation, and \(g_t\) is the gripper state. The incremental end-effector command
\(\delta x_t^{\mathrm{cmd}}\in\mathbb R^6\) is expressed in the configured frame relative to the preceding-step
TCP pose. Joint-correspondence commands from the isomorphic master are converted to slave TCP
increments before storage, thereby matching the keyboard interface. Across both interfaces, a
pipeline records synchronized state--action pairs for training at a nominal 15 Hz.

\subsubsection{Two-Stage Real-World Training}

VLA-Precision adopts two-stage post-training that combines demonstration-driven initialization with
real-world online improvement. Stage I fully fine-tunes pretrained \(\pi_{0.5}\) on task demonstrations
through flow-matching imitation learning. Stage II freezes the resulting base parameters
\((\Theta_f,\psi)\), initializes zero-effect trainable LoRA parameters \(\theta\) in the action expert,
updates only \(\theta\), and retains the corresponding frozen LoRA parameters \(\theta_0\) as the Stage-I
reference. The complete procedure is shown in Algorithm~\ref{alg:acob}.

\subsubsection{Robot Deployment}

To ensure safe, high-precision execution under physical contact, we implement robot-specific
Cartesian impedance controllers: force-mode wrench control for UR5e and torque-level control for Franka.

\textit{UR5e Force-Mode Impedance Control.} At control frequency \(f_c\), we compute the following
Cartesian PD wrench \(w_B\in\mathbb R^6\) and apply it through the UR \texttt{forceMode} interface:
\begin{equation}
w_B=
\begin{bmatrix}f_B\\ \mu_B\end{bmatrix}
=\begin{bmatrix}
K_p e_p-K_v\operatorname{sat}_{\delta_v}(v_t)\\
f_c^{-1}(K_R e_R-K_\omega\omega_t)
\end{bmatrix}.
\label{eq:ur_impedance}
\end{equation}
Here, \textcolor{black}{\(B\) denotes the reference frame.} \((p_t^d,R_t^d)\) and \((p_t,R_t)\) denote the desired and measured end-effector poses, and
\((v_t,\omega_t)\) \textcolor{black}{denote} the measured twist. With
\(\operatorname{sat}_{b}(x)=\operatorname{clip}(x,-b,b)\), the upper row converts position error
\(e_p=\operatorname{sat}_{\delta_p}(p_t^d-p_t)\) and linear velocity into force \(f_B\), whereas the
lower converts rotation error \(e_R=\operatorname{Log}(R_t^dR_t^\top)^\vee\) and angular velocity
into moment \(\mu_B\). \textcolor{black}{\(K_p,K_v\) and \(K_R,K_\omega\) denote translational and rotational stiffness and damping, respectively.}

\textit{\textcolor{black}{Franka Torque-Mode Impedance Control.}} We realize the commanded pose with the
following Cartesian impedance law, which produces the joint-torque command \(\tau_t\in\mathbb R^7\):
\begin{equation}
\begin{aligned}
\tau_t={}&(J_t^C)^\top(K_p e_t^C-K_d\nu_t^C)\\
&+N_t[K_N(q_0^J-q_t^J)-D_N\dot q_t^J]+\tau_C.
\end{aligned}
\label{eq:franka_torque}
\end{equation}
Here, \textcolor{black}{\(C\) denotes the reference frame, and \(J_t^C\) is the corresponding end-effector Jacobian.} \(e_t^C=\operatorname{sat}_{c}([(p_t^{d,C}-p_t^C)^\top,(e_{R,t}^C)^\top]^\top)\) and
\(\nu_t^C=[(v_t^C)^\top,(\omega_t^C)^\top]^\top\) denote the clipped pose error and twist in frame
\(C\). The first term maps the Cartesian PD wrench, with stiffness \(K_p\) and damping \(K_d\),
through \(J_t^C\) to the task torque. \textcolor{black}{\(K_N,D_N\) denote null-space stiffness and damping. \(q_0^J,q_t^J,\dot q_t^J\) denote reference joint positions, current joint positions, and velocities, respectively.} The second projects posture regulation toward \(q_0^J\) through
\(N_t=I-(J_t^C)^\top((J_t^C)^\top)^+\), stabilizing redundant joints without altering the end-effector
task. The final term \(\tau_C\) provides Polymetis Coriolis compensation. Together, these terms form
the final torque command \(\tau_t\) for all seven joints.

\section{Experiments}
\label{sec:experiments}

In this section, we first detail the experimental setup. Subsequently, we
evaluate the overall performance of VLA-Precision in real-world environments.
Furthermore, we evaluate the throughput and efficiency of the ACoB-Stream system architecture.
Finally, we validate the core mechanisms of the ACoB algorithm through ablation studies.

\subsection{Experimental Setup}
\label{sec:experimental_setup}

\begin{figure*}[!t]
    \centering
    \captionsetup[subfloat]{font=footnotesize,labelfont=rm,textfont=rm,justification=centering}
    \subfloat[\textcolor{black}{UR5e with PGI-140-80}]{%
        \includegraphics[width=0.24\textwidth]{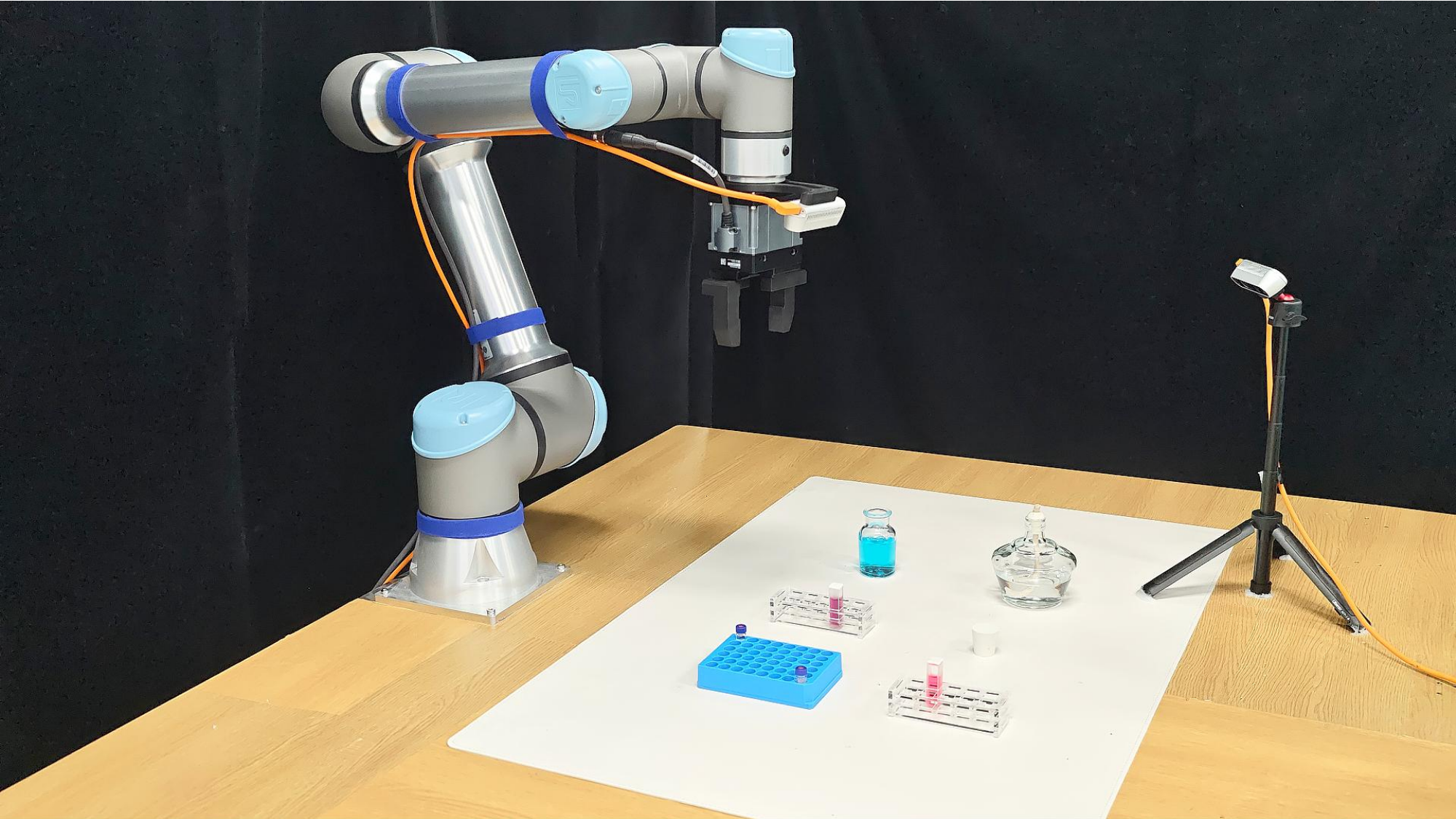}}
    \hfill
    \subfloat[\textcolor{black}{UR5e with LinkerHand L20}]{%
        \includegraphics[width=0.24\textwidth]{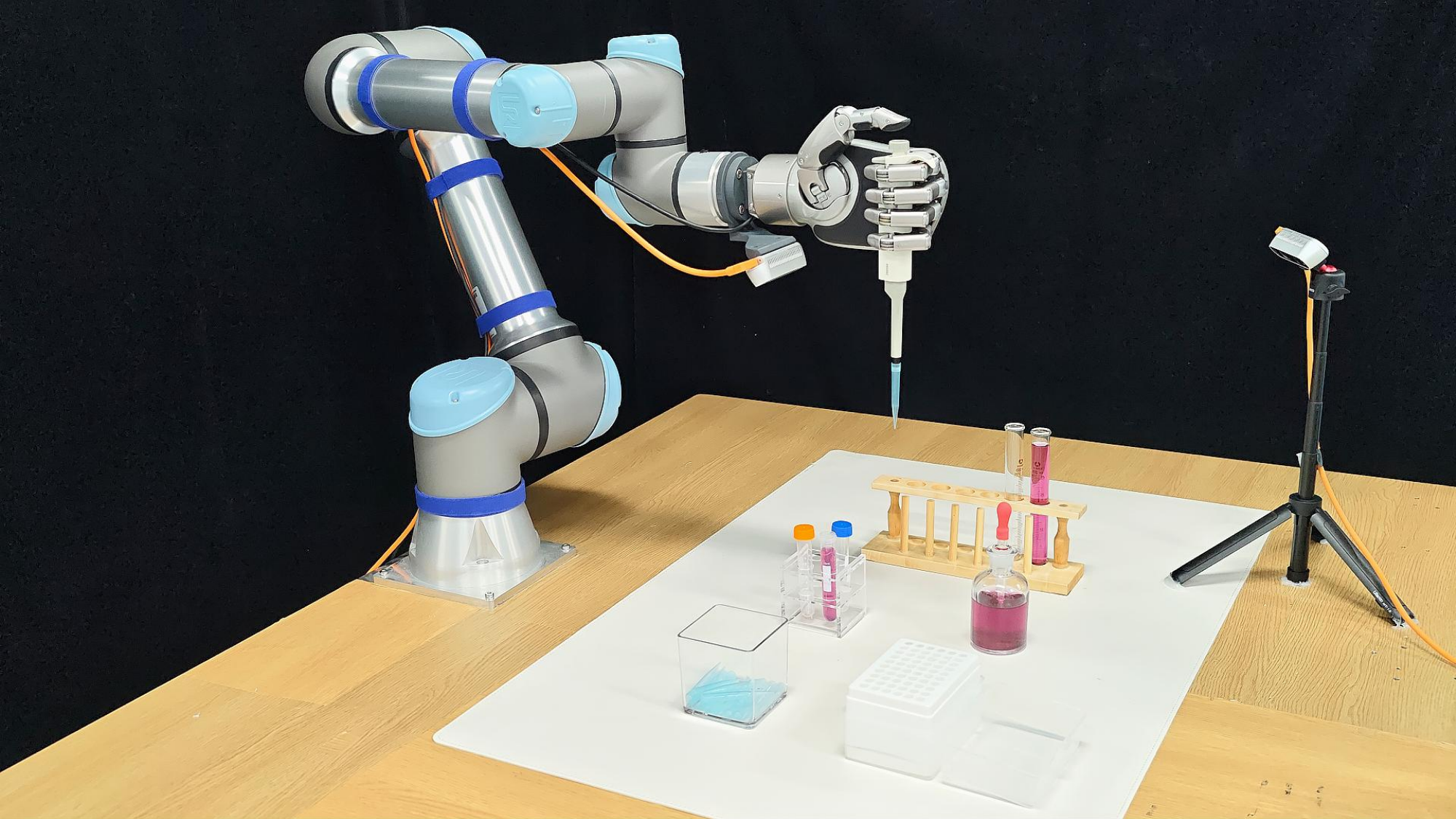}}
    \hfill
    \subfloat[\textcolor{black}{Franka with PGI-140-80}]{%
        \includegraphics[width=0.24\textwidth]{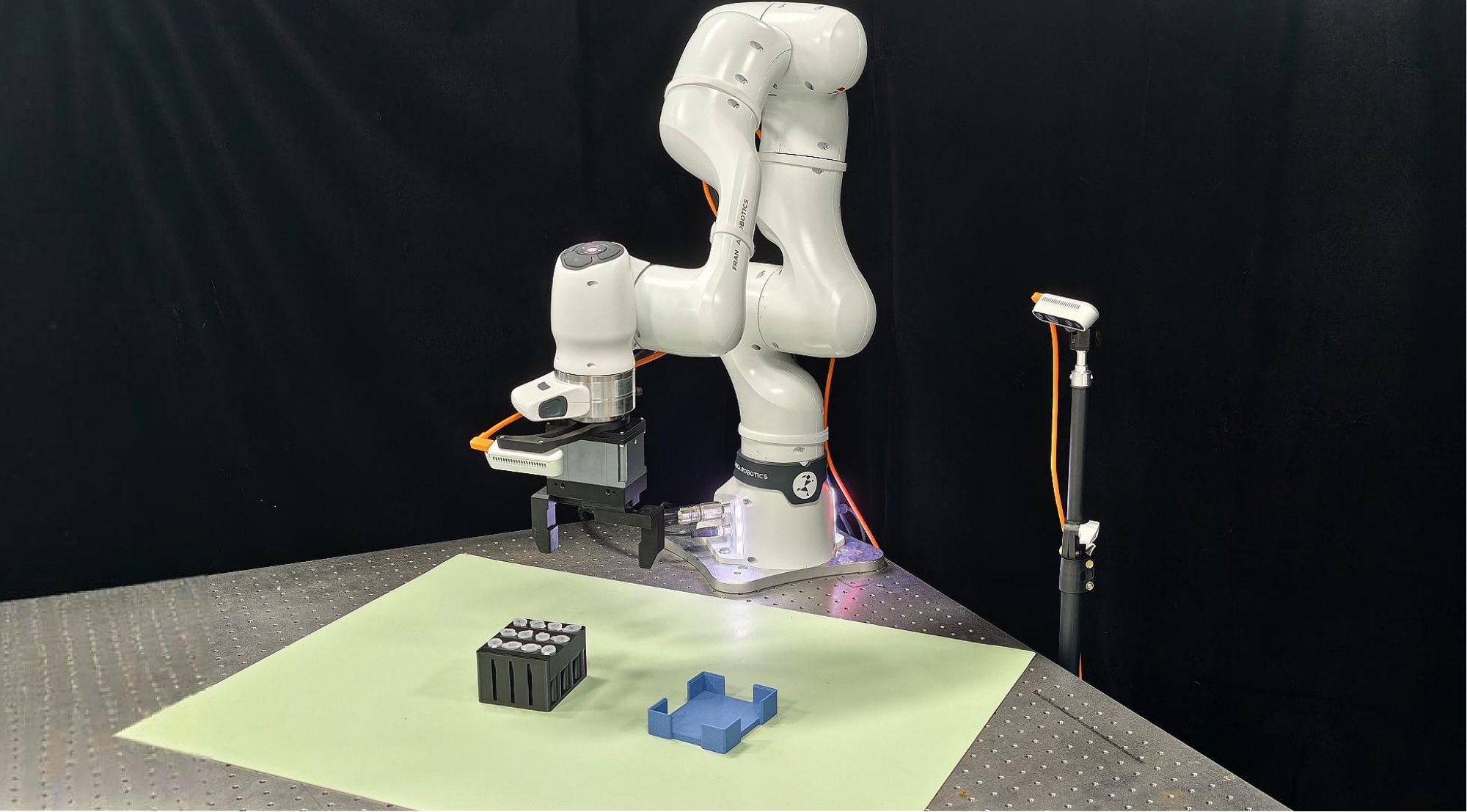}}
    \hfill
    \subfloat[\textcolor{black}{Dual UR5e with PGI-140-80}]{%
        \includegraphics[width=0.24\textwidth]{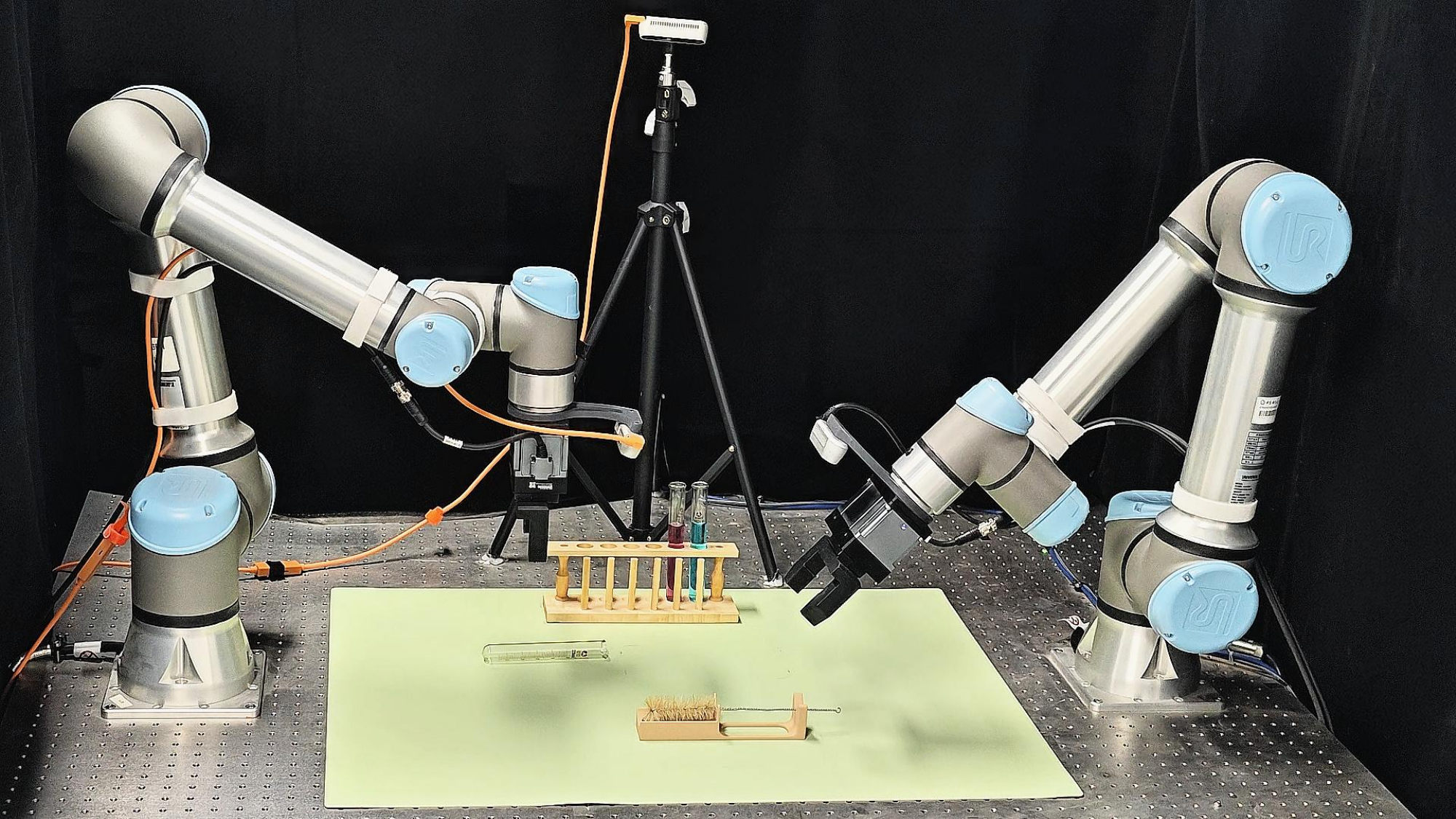}}
    \caption{Hardware platforms. Single-arm: one wrist camera and one external camera;
    dual-arm: two wrist cameras and one external camera.}
    \label{fig:robot_embodiments}
\end{figure*}

\textbf{Hardware Configuration.}
As shown in Fig.~\ref{fig:robot_embodiments}, our real-world experiments are
conducted on four robotic platforms: (1) a UR5e arm with a PGI-140-80
two-finger parallel gripper, (2) a UR5e arm with a LinkerHand L20 dexterous
hand, (3) two UR5e arms, each equipped with a PGI-140-80 gripper, and (4) a
Franka Research~3 arm with a PGI-140-80 gripper. Each single-arm platform is
equipped with one wrist camera and one external camera, whereas the dual-arm
platform is equipped with two wrist cameras and one external camera for
arm-local and shared scene views.

\begin{table*}[!t]
\caption{Real-world performance across chemistry manipulation tasks.}
\label{tab:baseline_task_results}
\centering
\scriptsize
\setlength{\tabcolsep}{1.3pt}
\renewcommand{\arraystretch}{1.1}
\resizebox{\textwidth}{!}{%
\begin{tabular}{@{}l@{\hspace{0.55em}}c@{\hspace{0.75em}}cccccc@{\hspace{0.75em}}cccccc@{}}
\toprule
\raisebox{-0.45\baselineskip}[0pt][0pt]{\textbf{Task}} & \raisebox{-17pt}[0pt][0pt]{\shortstack{\textbf{Ours}\\\textbf{Training}\\\textbf{(min)}}} & \multicolumn{6}{c}{\textbf{Success Rate (\%)}} & \multicolumn{6}{c}{\textbf{Average Episode Time (s)}} \\
\cmidrule(lr){3-8}\cmidrule(lr){9-14}
 &  & \raisebox{-0.6em}{\shortstack{HIL-\\SERL}} & ConRFT & \raisebox{-0.6em}{\shortstack{Robo-\\Dopamine}} & $\pi_0$ & $\pi_{0.5}$ & \textbf{Ours} & \raisebox{-0.6em}{\shortstack{HIL-\\SERL}} & ConRFT & \raisebox{-0.6em}{\shortstack{Robo-\\Dopamine}} & $\pi_0$ & $\pi_{0.5}$ & \textbf{Ours} \\[0.75ex]
\midrule
Tube Rack Loading & 27 & 0 & 5(+5\%) & 10(+10\%) & 75(+75\%) & 85(+85\%) & \textbf{100(+100\%)} & \textit{N/S} & 31.4(\ensuremath{\times}0.8) & 29.8(\ensuremath{\times}0.9) & 25.6(\ensuremath{\times}1.0) & 24.9(\ensuremath{\times}1.0) & \textbf{23.5(\ensuremath{\times}1.1)} \\
2 mL Vial Transfer & 34 & 0 & 5(+5\%) & 5(+5\%) & 80(+80\%) & 85(+85\%) & \textbf{100(+100\%)} & \textit{N/S} & 49.2(\ensuremath{\times}0.8) & 46.8(\ensuremath{\times}0.8) & 37.1(\ensuremath{\times}1.0) & 33.6(\ensuremath{\times}1.1) & \textbf{33.3(\ensuremath{\times}1.1)} \\
Cuvette Transfer & 65 & 0 & 0(+0\%) & 0(+0\%) & 35(+35\%) & 40(+40\%) & \textbf{90(+90\%)} & \textit{N/S} & \textit{N/S} & \textit{N/S} & 53.3(\ensuremath{\times}1.0) & 52.6(\ensuremath{\times}1.0) & \textbf{46.2(\ensuremath{\times}1.2)} \\
\midrule
Rubber Stopper Insertion & 52 & 5 & 10(+5\%) & 15(+10\%) & 50(+45\%) & 70(+65\%) & \textbf{100(+95\%)} & 44.6(\ensuremath{\times}0.6) & 38.9(\ensuremath{\times}0.7) & 37.7(\ensuremath{\times}0.7) & 28.1(\ensuremath{\times}1.0) & 23.8(\ensuremath{\times}1.2) & \textbf{19.2(\ensuremath{\times}1.5)} \\
Alcohol Lamp Extinguishing & 23 & 0 & 5(+5\%) & 10(+10\%) & 85(+85\%) & 90(+90\%) & \textbf{100(+100\%)} & \textit{N/S} & 29.4(\ensuremath{\times}0.8) & 27.8(\ensuremath{\times}0.8) & 22.4(\ensuremath{\times}1.0) & 21.1(\ensuremath{\times}1.1) & \textbf{21.0(\ensuremath{\times}1.1)} \\
Pipette Tip Attachment & 44 & 15 & 35(+20\%) & 40(+25\%) & 35(+20\%) & 50(+35\%) & \textbf{100(+85\%)} & 21.6(\ensuremath{\times}0.8) & 17.8(\ensuremath{\times}0.9) & 18.4(\ensuremath{\times}0.9) & 16.6(\ensuremath{\times}1.0) & 14.0(\ensuremath{\times}1.2) & \textbf{11.5(\ensuremath{\times}1.4)} \\
\midrule
Bulb Dropper Transfer & 28 & 5 & 10(+5\%) & 10(+5\%) & 90(+85\%) & 90(+85\%) & \textbf{100(+95\%)} & 32.7(\ensuremath{\times}0.8) & 30.8(\ensuremath{\times}0.9) & 29.6(\ensuremath{\times}0.9) & 26.4(\ensuremath{\times}1.0) & 24.7(\ensuremath{\times}1.1) & \textbf{24.4(\ensuremath{\times}1.1)} \\
Pipette Transfer and Ejection & 51 & 0 & 0(+0\%) & 0(+0\%) & 80(+80\%) & 90(+90\%) & \textbf{100(+100\%)} & \textit{N/S} & \textit{N/S} & \textit{N/S} & 36.7(\ensuremath{\times}1.0) & 35.4(\ensuremath{\times}1.0) & \textbf{33.6(\ensuremath{\times}1.1)} \\
\midrule
Tube Brushing & 87 & 0 & 0(+0\%) & 0(+0\%) & 5(+5\%) & 10(+10\%) & \textbf{95(+95\%)} & \textit{N/S} & \textit{N/S} & \textit{N/S} & 41.7(\ensuremath{\times}1.0) & 39.7(\ensuremath{\times}1.0) & \textbf{35.6(\ensuremath{\times}1.2)} \\
\midrule
\textbf{Average} & 45.8 & 2.8 & 7.8(+5\%) & 10(+7.2\%) & 59.4(+56.7\%) & 67.8(+65\%) & \textbf{98.3(+95.6\%)} & 50.4(\ensuremath{\times}0.6) & 45.3(\ensuremath{\times}0.7) & 44.5(\ensuremath{\times}0.7) & 32.0(\ensuremath{\times}1.0) & 30.0(\ensuremath{\times}1.1) & \textbf{27.6(\ensuremath{\times}1.2)} \\
\bottomrule
\end{tabular}
}
\end{table*}

\begin{table*}[!t]
\caption{\textcolor{black}{Experimental} settings for VLA-Precision and VLA baselines.}
\label{tab:experiment_settings}
\centering
\scriptsize
\setlength{\tabcolsep}{3.2pt}
\renewcommand{\arraystretch}{0.95}
\resizebox{0.95\textwidth}{!}{%
\begin{tabular}{@{}l@{\hspace{0.85em}}c@{\hspace{0.65em}}c@{\hspace{0.95em}}c@{\hspace{0.65em}}c@{\hspace{0.95em}}c@{\hspace{0.45em}}c@{\hspace{0.45em}}c@{}}
\toprule
\raisebox{-0.45\baselineskip}[0pt][0pt]{\textbf{Task}} & \multicolumn{2}{c}{$\pi_0$ / $\pi_{0.5}$} & \multicolumn{5}{c}{\textbf{Ours}} \\
\cmidrule(lr){2-3}\cmidrule(lr){4-8}
 & SFT Demos. & SFT Steps & Stage-I Demos. & Stage-I Steps & Max Horizon & Variation Range & \textcolor{black}{Alignment Tolerance} \\
\midrule
Tube Rack Loading & 120 & 25000 & 80 & 10000 & 150 & 40$\times$40 cm & \textcolor{black}{$\pm0.25$ mm / $\pm0.3^\circ$ ($z$)} \\
2 mL Vial Transfer & 150 & 25000 & 80 & 10000 & 250 & 40$\times$40 cm & \textcolor{black}{$\pm0.35$ mm} \\
Cuvette Transfer & 200 & 30000 & 120 & 15000 & 260 & 25$\times$25 cm & \textcolor{black}{$\pm0.3$ mm / $\pm2^\circ$ ($z$)} \\
\midrule
Rubber Stopper Insertion & 150 & 25000 & 80 & 5000 & 140 & 40$\times$40 cm & \textcolor{black}{$\pm0.5$ mm} \\
Alcohol Lamp Extinguishing & 120 & 25000 & 80 & 10000 & 120 & 40$\times$40 cm & \textcolor{black}{$\pm1$ mm} \\
Pipette Tip Attachment & 120 & 25000 & 60 & 5000 & 90 & 15 tip holes & \textcolor{black}{$\pm0.25$ mm} \\
\midrule
Bulb Dropper Transfer & 150 & 25000 & 80 & 15000 & 150 & 40$\times$20 cm & \textcolor{black}{$\pm6$ mm} \\
Pipette Transfer and Ejection & 150 & 25000 & 120 & 15000 & 220 & 2$\times$4 holes & \textcolor{black}{$\pm5$ mm} \\
\midrule
Tube Brushing & 200 & 30000 & 120 & 15000 & 250 & L: 10$\times$10 cm; R: 15$\times$10 cm & \textcolor{black}{$\pm3.25$ mm} \\
\bottomrule
\end{tabular}
}
\end{table*}

\begin{table}[!t]
\caption{Shared hyperparameters for real-world online training.}
\label{tab:shared_hyperparameters}
\centering
\scriptsize
\setlength{\tabcolsep}{5pt}
\renewcommand{\arraystretch}{1.08}
\begin{tabular}{@{}cc@{}}
\toprule
\textbf{Hyperparameter} & \textbf{Value} \\
\midrule
$\lambda_{\mathrm{rank}}$
    & 50 \\
($w_{\mathrm{BC}}$, $w_{\mathrm{rel}}$, $w_{\mathrm{ref}}$)
    & (0.25, 0.50, 0.25) \\
Initial reset range
    & $3\,\mathrm{cm}$ in $x$, $y$, and $z$ \\
Training GPUs
    & $4\times$ NVIDIA A800 GPUs \\
\bottomrule
\end{tabular}
\end{table}

\textbf{Baselines.}
We compare VLA-Precision with three real-world RL methods and two large VLA
baselines. HIL-SERL~\cite{luo2024hilserl} combines high-update-to-data
off-policy RL with human corrections and reports near-perfect success on
dexterous tasks after 1--2.5 hours of real-world training.
ConRFT~\cite{chen2025conrft} fine-tunes the Octo VLA through
offline-to-online RL with a consistency-policy action head, reporting 96.3\%
average success across eight real-world tasks within 45--90 min.
Robo-Dopamine~\cite{tan2025robodopamine} performs RL on the Octo
VLA using dense rewards generated by a process reward model adapted with one
demonstration, reporting 95.2\% average success after 150 real-world rollouts.
$\pi_0$~\cite{black2025pi0} and $\pi_{0.5}$~\cite{black2025pi05} are
pretrained flow-based large VLAs that we fine-tune on each task without
Stage-II online RL.

\textbf{Tasks and Metrics.}
We design nine high-precision manipulation tasks in chemistry laboratories
(Fig.~\ref{fig:additional_realworld_examples}) to evaluate performance across four task categories: contact-rich,
contact-light, contact-free, and bimanual coordination.
\textcolor{black}{To quantify the precision challenges of these tasks, Table~\ref{tab:experiment_settings} reports positional and angular tolerances, defined as the maximum allowable lateral displacement perpendicular to the insertion direction and angular deviation about the specified TCP-frame axis.}
To rigorously evaluate generalization, we deliberately vary object poses and
initial arm poses in every task, with the evaluated ranges for object poses and
arm resets reported in the Variation Range column of
Table~\ref{tab:experiment_settings} and Table~\ref{tab:shared_hyperparameters}, respectively.

\begin{figure*}[t]
    \centering
    \includegraphics[width=0.98\textwidth]{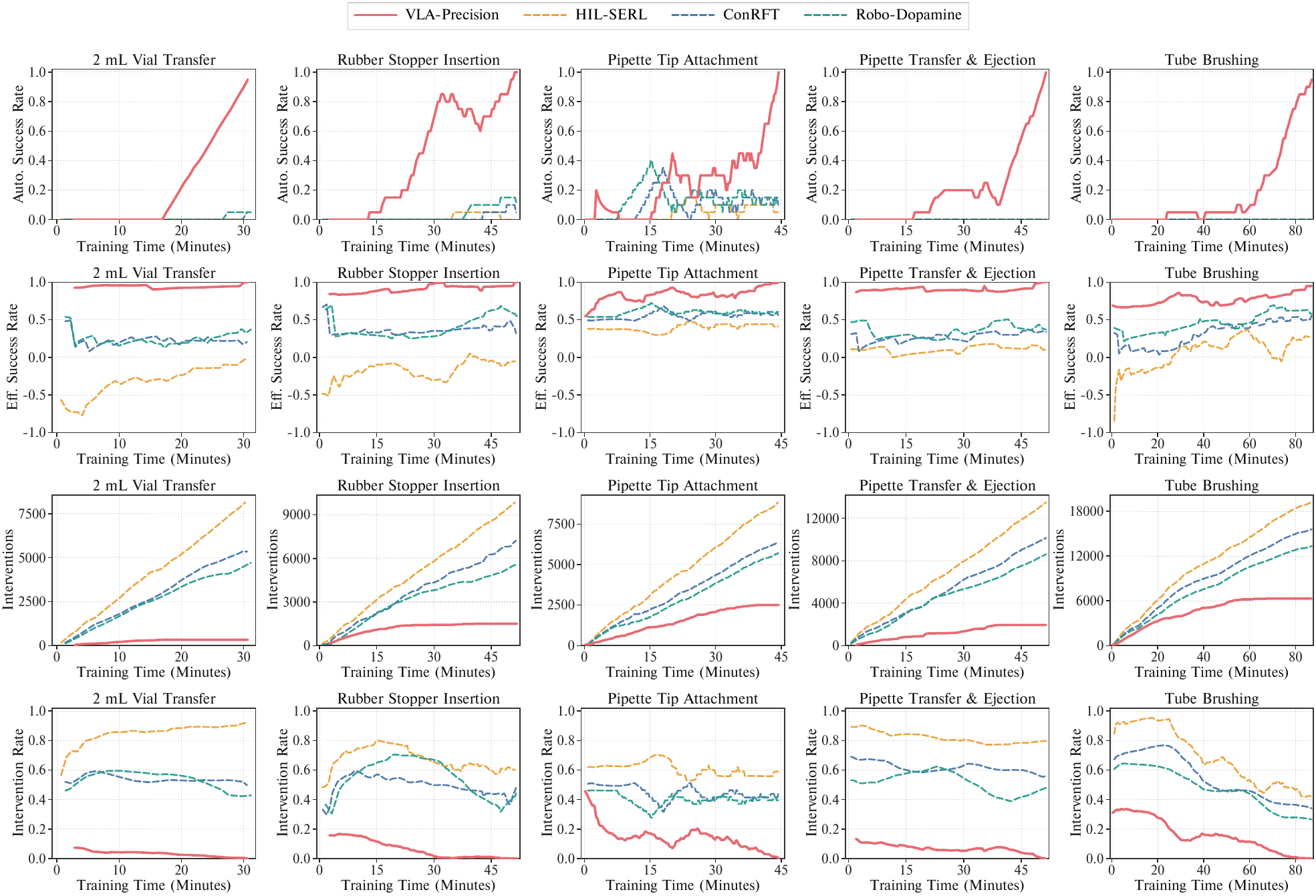}
    \caption{Real-world online RL training dynamics on five representative tasks.
    Autonomous success rate is the 20-episode rolling mean of autonomous task
    completion. Effective success rate is the 20-episode rolling mean of task
    completion, including assisted completion, minus the intervention rate.
    Interventions denote cumulative human-corrected low-level action steps, and
    intervention rate is the 20-episode rolling mean of their per-episode fraction.}
    \label{fig:vla_precision_training}
\end{figure*}

\smallskip
\noindent\textit{1) High-Precision Contact-Rich Tasks.}
\textcolor{black}{These tasks require submillimeter positional precision ($\pm0.25$ to $\pm0.35$ mm) and tight angular alignment (down to $\pm0.3^\circ$), with small deviations potentially causing jamming or damage during insertion.}

\textbf{Cuvette Transfer.} The robot sequentially transfers two cuvettes from
one rack to two designated diagonal slots in another rack. The fully transparent
glass cuvettes and racks complicate visual perception, while their fragility
demands precise pose and force control.

\textbf{2 mL Vial Transfer.} The robot sequentially transfers two 2 mL vials
from one rack to diagonal slots in another. The small vial diameter imposes
tight alignment tolerances.

\textbf{Tube Rack Loading.} The robot lifts a centrifuge-tube rack and inserts
it into a loading slot. Insertion requires complete pose alignment with the
slot and precise force control.

\smallskip
\noindent\textit{2) High-Precision Contact-Light Tasks.}
\textcolor{black}{These tasks require precise positional alignment ($\pm0.25$ to $\pm1$ mm) under limited and intermittent contact, with small alignment errors or excessive contact forces potentially destabilizing the objects.}

\textbf{Rubber Stopper Insertion.} The robot inserts a rubber stopper into a
reagent-bottle opening. A high grasp can crush the stopper, whereas a low grasp
leaves insufficient insertion clearance, and bottle motion demands stable
control.

\textbf{Alcohol Lamp Extinguishing.} The robot covers an alcohol-lamp wick
with an extinguisher cap. Visual interference from the flame demands precise
action generation through fused visual and force feedback.

\textbf{Pipette Tip Attachment.} The robot aligns a pipette with a tip and
presses it into place. The narrow tip aperture requires precise
alignment and appropriate axial force.

\smallskip
\noindent\textit{3) High-Precision Contact-Free Tasks.}
\textcolor{black}{These tasks require positional alignment ($\pm5$ to $\pm6$ mm) under visual--proprioceptive inference, with accumulated errors potentially causing rim collisions or spillage without contact feedback.}

\textbf{Bulb Dropper Transfer.} The robot uses a flexible bulb dropper to
aspirate and dispense liquid. Visual guidance must coordinate finger actuation
with precise end-effector alignment, leaving little tolerance for timing or
pose errors.

\textbf{Pipette Transfer and Ejection.} The robot transfers liquid and ejects
the used pipette tip. The long sequence
requires visually guided coordination of hand state and end-effector motion
while limiting accumulated error.

\smallskip
\noindent\textit{4) High-Precision Bimanual Coordination Task.}
\textcolor{black}{This task requires precise alignment through bimanual coordination ($\pm3.25$ mm), with relative positioning errors potentially causing edge collisions that obstruct insertion.}

\begin{figure*}[t]
    \centering
    \includegraphics[width=\textwidth]{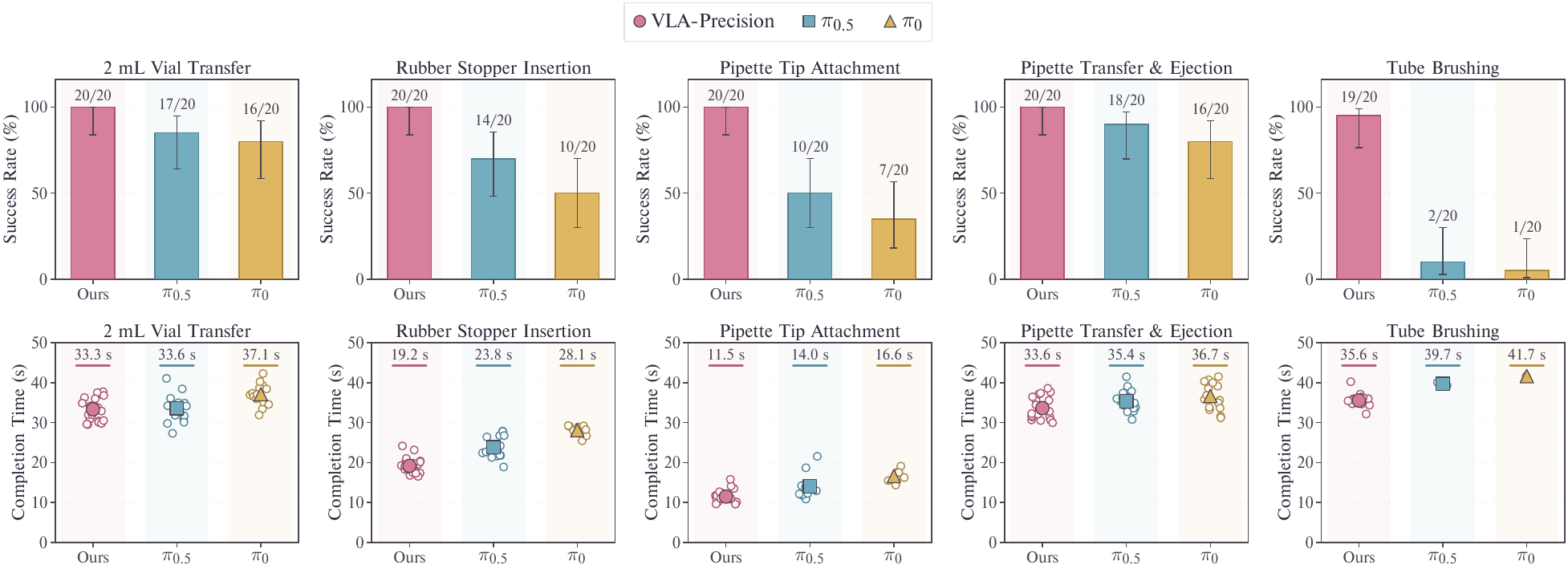}
    \caption{Real-world test performance on five representative tasks. Each
    method--task pair comprises 20 trials. Top: success rates with two-sided
    Wilson 95\% confidence intervals. Bottom: successful-trial completion times
    (open circles) and their means (filled markers).}
    \label{fig:vla_precision_test}
\end{figure*}

\textbf{Tube Brushing.} One arm stabilizes a test tube while the other inserts
and manipulates a brush. The transparent tube complicates visual localization,
while brushing requires high-precision coordination between the two arms.

\subsection{Real-World Evaluation of VLA-Precision}
\label{sec:vla_precision_real_world}

\textbf{Overall Performance.}
Table~\ref{tab:baseline_task_results} reports the real-world experimental
results across nine chemistry manipulation tasks. Compared with the baselines,
VLA-Precision achieves the highest average success rate of 98.3\% after an
average of 45.8 min of online training per task and completes successful trials
in 27.6 s on average. It succeeds in 177 of 180 held-out trials, reaches 100\%
on seven tasks, and maintains at least 90\% success on every task. Relative to
$\pi_{0.5}$ and $\pi_0$, VLA-Precision improves average success by 30.5\% and
38.9\% and improves average execution speed by 8.7\% and 15.9\%, respectively.
Its worst-task success rate is 90\%, compared with 10\% for $\pi_{0.5}$ and
5\% for $\pi_0$. Relative to the strongest real-world RL baseline,
Robo-Dopamine, VLA-Precision improves average success by 88.3\% and average
execution speed by 61.2\%.

The results reveal the limitations of the baselines. HIL-SERL
learns a compact task-specific policy but starts online RL with limited
task-specific competence. Although effective in shorter-horizon settings
with limited task variation, it reaches only 2.8\% average success on our
longer-horizon, pose-randomized suite. ConRFT and Robo-Dopamine demonstrate
online RL on the Octo VLA, with Robo-Dopamine
providing dense progress rewards. Although both enter online RL after task-specific offline training on Octo,
direct maximization of absolute Q estimates remains
vulnerable to optimistic value errors and policy drift. Their average success
is limited to 7.8\% and 10.0\%. Full-parameter fine-tuning of $\pi_0$ and
$\pi_{0.5}$ retains stronger pretrained VLA priors and improves average success
to 59.4\% and 67.8\%, but \textcolor{black}{demonstration-only adaptation remains insufficient for precise manipulation across these tasks}.
The 98.3\% average success of VLA-Precision shows that ACoB \textcolor{black}{can further
optimize task-finetuned large VLAs through asymmetric co-bootstrapping to achieve
precise manipulation under tight tolerances}.

\textbf{Online RL Training Performance.}
Fig.~\ref{fig:vla_precision_training} shows four online RL metrics on five
representative tasks. At the final task-specific comparison horizons,
VLA-Precision leads all four metrics. Across the five tasks, autonomous success
rate, effective success rate, interventions, and intervention rate average
98.0\%, 99.0\%, 2,517.4, and 0.024\%, respectively. \textcolor{black}{Relative to Robo-Dopamine,
the strongest baseline, VLA-Precision raises autonomous and effective success rates by 92.0 and
49.5 percentage points while reducing the rate of human intervention by 40.46 percentage points.}

For the high-precision contact-rich 2 mL vial transfer task, successive
pick-and-place operations form a multistep sequence. HIL-SERL begins with
limited task-specific competence and never achieves autonomous success. Its
effective success rate remains below zero because assisted success cannot offset
the 91.95\% final intervention rate. ConRFT and Robo-Dopamine enter with
task-specific behavior learned offline and lower intervention rates to 49.70\%
and 42.84\%, yet both end at only 5\% autonomous success. Their BC
regularization anchors the actor, but direct absolute-Q maximization remains
vulnerable to optimistic value errors and policy drift, preventing reliable
autonomous multistep behavior.

For the high-precision contact-light rubber-stopper insertion and pipette-tip
attachment tasks, HIL-SERL ends stopper insertion at 0\% autonomous success and
the 60.07\% intervention rate. ConRFT and Robo-Dopamine end at only 5\% and 10\%,
with rates of 48.08\% and 45.63\%, respectively. Offline-learned task behavior
reduces intervention for the latter two, but unstable updates perturb the
compliant stopper and easily displaced bottle, preventing stable behavior. Tip
attachment is the only short-sequence task without gripper control. The three
baselines peak at 15\%, 35\%, and 40\% autonomous success but fall to 5\%, 10\%,
and 15\% across 15 randomized holes. The higher VLA peaks show faster local
acquisition, but susceptibility to optimistic value errors and policy drift
prevents robust pose generalization.

\begin{figure}[t]
    \centering
    \includegraphics[width=\columnwidth]{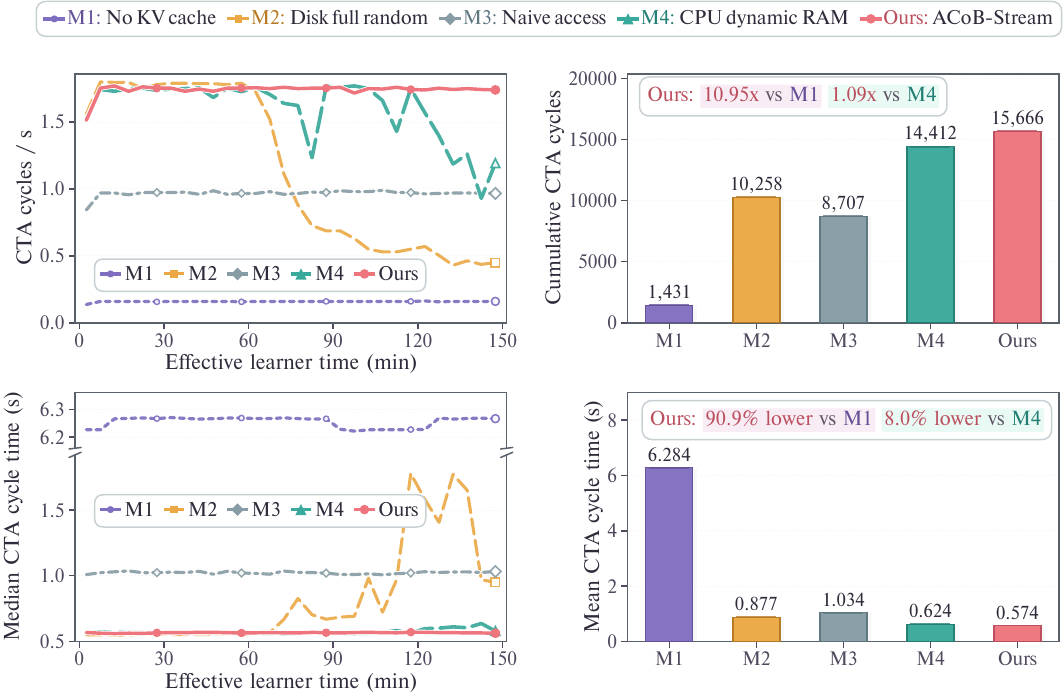}
    \caption{Training throughput and latency comparison. Top left: CTA throughput versus
    effective learner time. Top right: cumulative CTA cycles after 150 min. Bottom left:
    median CTA cycle time versus effective learner time. Bottom right: mean CTA cycle time.
    A CTA cycle comprises one critic-only update and one joint critic--actor update.}
    \label{fig:acob_stream_benchmark}
\end{figure}

For the high-precision contact-free pipette transfer and tip ejection task,
the alignment demand of tip attachment expands into a long sequence. All three
baselines remain at 0\% autonomous success. Starting with weaker task-specific
competence, HIL-SERL accumulates 13,512 interventions with a 79.82\% intervention
rate, while ConRFT and Robo-Dopamine require 10,154 and 8,614 interventions at
rates of 56.10\% and 47.96\%, respectively. Stable insertion into and withdrawal
from narrow tube openings require precise actions conditioned on vision and
robot state. Task-specific offline initialization reduces intervention, and
Robo-Dopamine adds dense progress rewards. However, direct absolute-Q
maximization leaves both methods vulnerable to accumulated value and motion
errors, so alignment cannot remain stable over the full sequence.

For the bimanual high-precision test-tube brushing task, the coupled dual-arm
data distribution and long sequence leave all baselines at 0\% autonomous
success. Starting with weaker task-specific competence, HIL-SERL accumulates
19,196 interventions. ConRFT and Robo-Dopamine begin with task behavior learned
offline and reduce the counts to 15,587 and 13,322, yet neither succeeds
autonomously. Their BC regularization anchors the actor, but absolute-Q policy
drift compounds across both arms and destabilizes the sequence.

ACoB instead rapidly absorbs online corrections through intervention-guided behavior
cloning, continually calibrates value estimates through global returns and
state-matched preferences, and stabilizes relative-advantage updates with a
frozen reference. Together, these mechanisms raise the autonomous success rate and
reduce the intervention rate across all five tasks, validating asymmetric
co-bootstrapping.

\textbf{Offline VLA Test Performance.}
We further evaluate VLA-Precision, $\pi_{0.5}$, and $\pi_0$ offline on the same
five representative tasks. As summarized
in Table~\ref{tab:experiment_settings}, $\pi_0$ and $\pi_{0.5}$ receive 120--200
demonstrations and 25{,}000--30{,}000 SFT steps, whereas VLA-Precision uses
60--120 demonstrations and 5{,}000--15{,}000 Stage-I steps before online RL.

Fig.~\ref{fig:vla_precision_test} reports 20 trials per method and task.
Overall, VLA-Precision succeeds in 99 of 100 trials, compared with 61 for
$\pi_{0.5}$ and 50 for $\pi_0$, improving average success by 38 and 49
percentage points, respectively. Notably, it reaches 100\% on all four
single-arm tasks and 95\% on bimanual tube brushing. Across the single-arm
tasks, $\pi_{0.5}$ ranges from 50--90\% and $\pi_0$ from 35--80\%. On brushing,
they fall to 10\% and 5\%. VLA-Precision also completes successful trials in
26.65 s on average, compared with 29.32 s for $\pi_{0.5}$ and 32.05 s for
$\pi_0$.

More importantly, the task-level gaps expose the limits of demonstration-only
fine-tuning under pose randomization, precision demands, and long-horizon
coordination. Collectively, these results show that online RL strengthens the
task-specific VLA policy beyond demonstration-only adaptation, yielding
\textcolor{black}{higher success rates} across the five tasks.

\subsection{Throughput and Efficiency Evaluation of ACoB-Stream}
\label{sec:acob_stream_systems}

To quantify the system-level advantages of ACoB-Stream in large-VLA online RL,
we compare it with four actor--learner system baselines in online RL update
throughput and computational efficiency. These baselines extend the
asynchronous actor--learner architecture. HIL-SERL~\cite{luo2024hilserl} is
the most representative real-world implementation of this architecture. It
decouples robot interaction from policy optimization so that data collection
and policy updates proceed without synchronous blocking. We adapt this
architecture to large VLAs and construct three system-level ablations
plus a competitive CPU-resident alternative.

The first three baselines selectively remove ACoB-Stream mechanisms.
\textbf{M1: No KV Cache} removes cross-update reuse of frozen-prefix contexts
and recomputes the required contexts during every update. \textbf{M2: Disk
Full Random} retains the disk-backed context store but replaces sliding-window
sampling with uniform sampling over the full replay history. \textbf{M3: Naive
Context Access} removes objective-aligned context retrieval by always loading
both current and successor contexts, and disables on-demand context prefetch.
\textbf{M4: Dynamic CPU RAM} is a competitive alternative
that maintains a dynamically sized replay and KV active window in learner CPU
RAM, applies FIFO eviction, and retains disk copies for persistence.
\textbf{\textcolor{black}{Ours: ACoB-Stream}} is the full system, combining cross-update reuse
of frozen-prefix contexts, context deduplication with disk-backed sliding-window
sampling, and objective-aligned context retrieval with on-demand context
prefetch.

\begin{figure*}[t]
    \centering
    \includegraphics[width=0.98\textwidth]{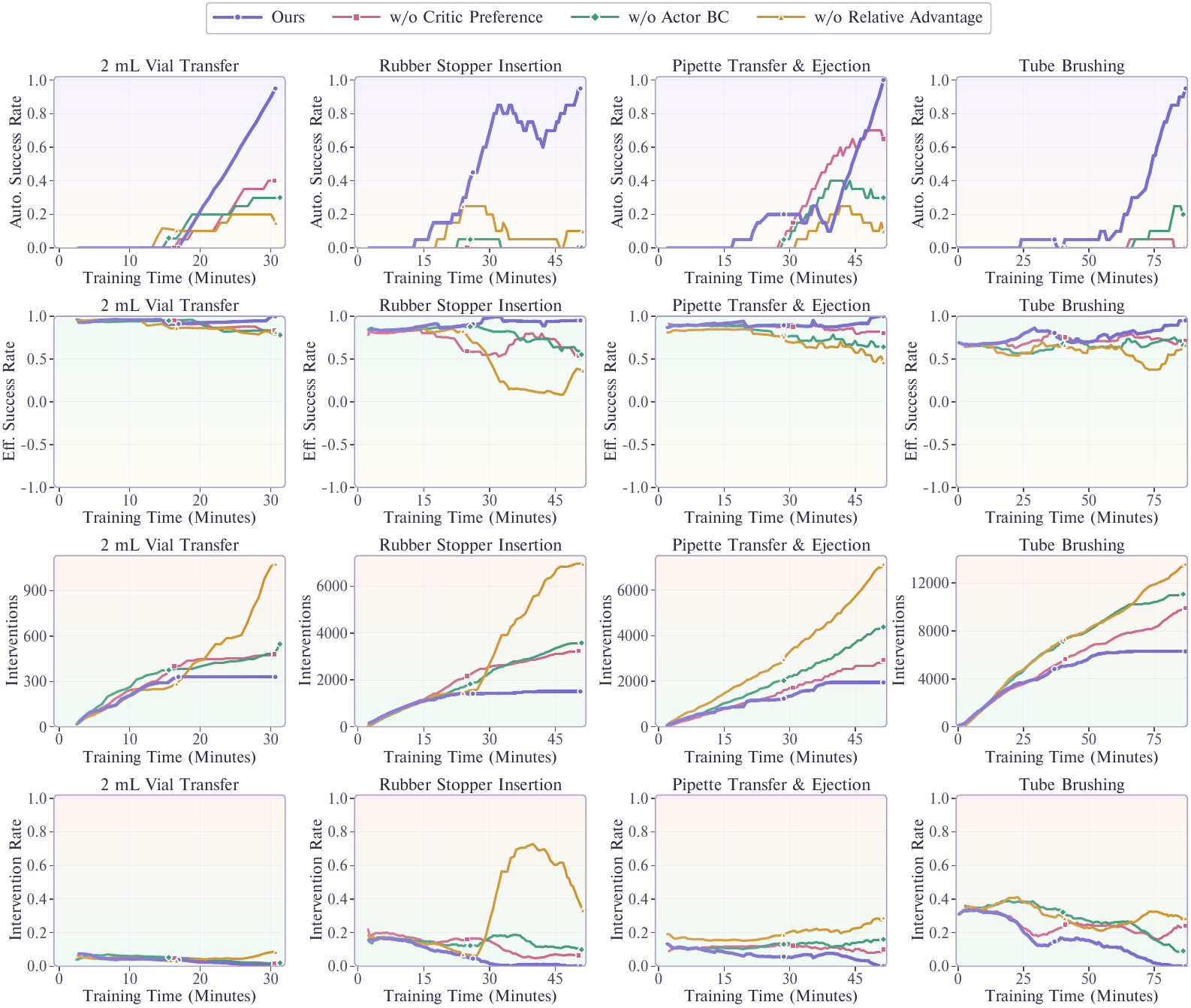}
    \caption{ACoB ablation training dynamics across four representative tasks. Columns
    denote tasks; rows show autonomous success rate, effective success rate, cumulative
    interventions, and intervention rate over training time. Rate metrics are 20-episode
    rolling means.}
    \label{fig:vla_precision_ablation}
\end{figure*}

Fig.~\ref{fig:acob_stream_benchmark} reports the system evaluation.
Disk ACoB-Stream achieves the highest update throughput and the
lowest mean critic-to-actor (CTA) cycle latency, completing 15,666 CTA cycles in 150 min at
1.7407 CTA/s and 0.574 s per cycle. Compared with No KV Cache, it
delivers 10.95$\times$ the throughput and reduces mean latency by 90.9\%,
demonstrating the computational value of cross-update frozen-prefix reuse.
Against Disk Full Random, it provides $1.53\times$ the throughput and
$34.5\%$ lower mean latency. Moreover, full-history random sampling
begins to degrade at 43~min as the replay working set exceeds the Linux
page-cache capacity, causing more frequent disk-to-CPU transfers, with
its throughput eventually declining from 1.56 to 0.45~CTA/s, while Disk
ACoB-Stream reaches 1.74~CTA/s in the final 5-min bin, showing that
sliding-window sampling preserves disk locality as replay grows.
Objective-aligned retrieval and on-demand prefetch jointly provide
1.80$\times$ the throughput and 44.4\% lower mean latency than Naive Context
Access, confirming the benefit of loading and transferring only required
contexts. Although Dynamic CPU RAM remains competitive, it requires a more
complex memory hierarchy to manage dynamically sized CPU-resident replay and
KV windows, coordinate FIFO eviction, and maintain disk persistence. Disk
ACoB-Stream delivers 1.09$\times$ the throughput and 8.0\% lower
latency while eliminating CPU--disk active-window coordination overhead.

\subsection{Ablation Experiments for the ACoB Algorithm}
\label{sec:ablation}

To evaluate the contributions of the core mechanisms in ACoB, we compare
the full algorithm with three ablated variants on four representative tasks.
\textbf{w/o critic preference} removes the intervention-guided critic
preference loss and computes the same paired action comparison through equivalent
relative Q-value differences rather than relative advantage differences.
\textbf{w/o actor BC} removes intervention-guided behavior cloning and
renormalizes the remaining actor-loss weights. \textbf{w/o relative advantage}
replaces relative-advantage policy improvement with direct sampled-action Q
maximization.

\textbf{Overall Results.}
Fig.~\ref{fig:vla_precision_ablation} reports results at the per-task horizons
shared by all four variants. Full ACoB reaches 96.25\% final autonomous success
with a 0.03\% intervention rate on average. Removing critic preference, actor
BC, or relative advantage reduces autonomous success to 26.25\%, 20.00\%, and
8.75\%, while increasing intervention rates to 10.57\%, 9.24\%, and 24.93\%.
Interventions across the four tasks also rise from 10,080 to 16,580, 19,581,
and 28,806. These degradations demonstrate that the evaluated mechanisms
are required for reliable autonomous improvement.

\begin{figure}[!t]
    \centering
    \includegraphics[width=\columnwidth]{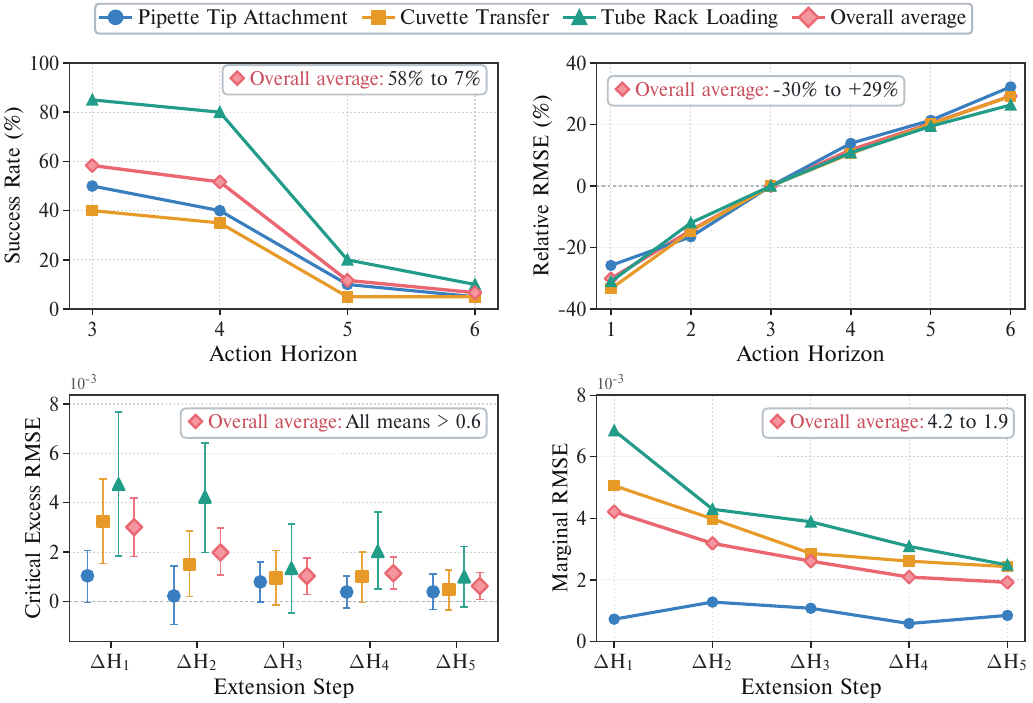}
    \caption{Action-horizon analysis \textcolor{black}{of $\pi_{0.5}$} on three representative precision-insertion
    tasks. Top left: autonomous success rate versus executed action horizon $H_e$.
    Top right: relative action-prediction RMSE, measuring predicted--demonstration
    action discrepancy as the percentage change from $H_e=3$. Bottom left: excess
    marginal RMSE in interaction-critical phases relative to normal motion. Bottom
    right: marginal RMSE increase for each extension $\Delta H_i$, where $\Delta H_i$
    denotes increasing $H_e$ from $i$ to $i+1$. Colored markers denote task means,
    red diamonds their equal-task-weighted mean, and bottom-left whiskers 95\%
    episode-bootstrap confidence intervals.}
    \label{fig:action_horizon_analysis}
\end{figure}

\textbf{w/o critic preference.}
Without state-matched preference ranking, TD observes only executed actions and
cannot correct the value of overwritten policy proposals. Returns obtained after
human correction can therefore leave failed proposals overvalued, weakening
subsequent action comparisons. The 70.00-point success drop demonstrates that
critic preference resolves this local credit ambiguity and progressively
calibrates action-dependent value estimates.

\textbf{w/o actor BC.}
Removing actor BC prevents successful and human-corrected actions from being
directly absorbed by the action expert. Behavioral improvement must then rely
on slower, indirect critic guidance, increasing intervention and degrading the
online experience underlying value calibration. The 76.25-point success drop
confirms that actor BC rapidly converts corrections into stable behavior and
improves the experience supporting subsequent value learning.

\textbf{w/o relative advantage.}
Replacing relative-advantage improvement with direct Q maximization produces
the largest degradation: an 87.50-point success drop and 24.93\% intervention
rate. Direct maximization exposes the action expert to absolute value scale and
optimistic critic errors, favoring high predicted Q values over reliable local
improvement. Relative advantage is therefore essential for translating value
estimates into stable policy improvement while suppressing policy drift.

In summary, the \textcolor{black}{ablation experiments} validate asymmetric co-bootstrapping in ACoB: actor BC rapidly
improves behavior and online experience quality, while critic preference
progressively calibrates value estimates, yielding \textcolor{black}{relative advantages
that stabilize policy improvement, enabling efficient learning to perform
tight-tolerance precise manipulation}.

\section{Discussion}

Our results indicate that reliable real-world RL of large VLAs depends on
resolving an early temporal asymmetry: task fine-tuning yields a capable policy
before online experience can adequately calibrate the critic. Consequently,
conservative value-guided updates learn slowly, whereas aggressive updates risk
policy drift. ACoB \textcolor{black}{mitigates this asymmetry} by rapidly improving behavior from successful
and corrected actions while progressively calibrating value estimates, \textcolor{black}{enabling
efficient policy improvement toward precise manipulation under tight positional and angular tolerances}.

Practical real-world RL of large VLAs also depends on how efficiently new
experience is converted into policy updates. As data accumulate, redundant
computation, transfer, and random access increase update latency. ACoB-Stream
shows that jointly managing reusable computation, experience, and policy state
can \textcolor{black}{sustain high update throughput} under finite compute and memory during
continuous online training.

More broadly, our findings show that real-world RL can improve an already
capable VLA while preserving its task-specific competence. In high-precision
tasks, small value-induced policy shifts can \textcolor{black}{disrupt policy improvement under tight task tolerances}.
Maintaining behavioral stability during improvement is therefore central to
\textcolor{black}{learning precise manipulation efficiently through online RL}.

Building on these findings, we discuss two broader research directions that warrant further investigation:

\textit{1) Long-Horizon Bimanual Real-World RL for VLAs.}
The evaluation includes one bimanual precision task but not longer sequences
with multiple precision-critical stages, where local errors accumulate and
degrade later operations. Longer sequences make the reference scheme for delta
actions increasingly important. Let $x_t^{\mathrm{ref}}$ denote the TCP pose at
the start of the current action horizon. Step-wise and chunk-wise forms decode
as
\begin{equation}
\begin{gathered}
\widetilde x_{t+k}^{\mathrm{step}}
=x_t^{\mathrm{ref}}+\sum_{j=1}^{k}z_{t,j}^{\mathrm{step}},\\
\widetilde x_{t+k}^{\mathrm{chunk}}
=x_t^{\mathrm{ref}}+z_{t,k}^{\mathrm{chunk}},\quad k=1,\ldots,H_e,
\end{gathered}
\label{eq:action_delta_formats}
\end{equation}
where step-wise deltas reference the preceding TCP pose, keeping the target
distribution compact and bounded but recursively accumulating prediction
errors. Chunk-wise deltas reference the horizon-start pose and avoid recursive
accumulation, although their target distribution broadens as $H_e$ increases.
We therefore use step-wise deltas with $H_e=3$ to preserve frequent closed-loop
correction while limiting recursive accumulation in the current precision
setting.

\textcolor{black}{To examine the effect of increasing the execution horizon}, we evaluate
\textcolor{black}{the $\pi_{0.5}$ fine-tuned on task demonstrations using} step-wise deltas at different executed action horizons on three representative
high-precision insertion tasks, as shown in
Fig.~\ref{fig:action_horizon_analysis}. Specifically, increasing $H_e$ from 3 to 6 reduces
average autonomous success from 58.3\% to 6.7\% and increases relative
action-prediction RMSE by 29.3\%. Although the marginal RMSE introduced by
each horizon extension decreases, it remains positive and is consistently
higher during interaction-critical phases than during normal motion.
These results expose the limitation of step-wise deltas as the execution
horizon grows and motivate chunk-wise deltas for long-horizon bimanual
manipulation. Feng et al.~\cite{feng2026demystifying} provide complementary
theoretical and empirical support by deriving $O(H_e)$ and $O(1)$ worst-case
\textcolor{black}{decoding-error amplification factors} for step-wise and chunk-wise deltas, respectively, and reporting
superior chunk-wise performance \textcolor{black}{in their controlled real-world imitation-learning evaluations}.

\textit{2) Multi-Task Real-World RL for Large VLAs.}
An important extension beyond current task-specific training is to apply ACoB
to joint multi-task optimization of a shared large VLA, enabling interaction
experience to transfer across related skills while reducing per-task data
collection. Task-conditioned value learning with balanced interaction data
and retained demonstrations provides a feasible path to limiting interference
and catastrophic forgetting, as supported by recent multi-task real-world VLA
post-training~\cite{pan2026sop}.

\section{Conclusion}

In this work, we present VLA-Precision, an efficient framework for real-world
online RL of large VLAs \textcolor{black}{to enable high-precision manipulation under tight tolerances}.
Within this framework, ACoB couples rapid behavioral
learning with progressive value calibration across timescales, \textcolor{black}{suppressing policy drift for stable and efficient policy optimization}. To make this optimization efficient for large VLAs, ACoB-Stream
establishes invariant-state decoupling and on-demand streaming as design
principles, delivering up to 10.9$\times$ improvements in throughput and
computational efficiency. Extensive evaluations across nine high-precision
\textcolor{black}{chemistry manipulation tasks} and four \textcolor{black}{robotic platforms} demonstrate 98.3\% average success
in 45.8 min/task, with 27.6 s episodes running at 1.2$\times$ and 1.8$\times$
the speeds of VLA and RL baselines.

\bibliographystyle{ieeetr}
\bibliography{main}

\end{document}